\documentclass[twoside,11pt]{article}
\usepackage{jmlr2e}

\usepackage[utf8]{inputenc}
\usepackage{graphicx}
\usepackage{amsmath}
\usepackage{algorithm2e}
\usepackage{tikz} 
\usetikzlibrary{positioning,arrows.meta,quotes,bayesnet,matrix}
\usepackage{color}
\usepackage{caption}
\usepackage{subcaption}
\usepackage{capt-of}
\usepackage{bbm}
\usepackage{amssymb}
\usepackage{stackrel}
\usepackage{pifont}
\newcommand{\paren}[1]{\mathopen{}\left({#1}_{{}_{}}\,\negthickspace\right)\mathclose{}}
\newcommand{\bracket}[1]{\mathopen{}\left[ {#1}_{{}_{}}\,\negthickspace\right]\mathclose{}}
\newcommand{\set}[1]{\mathopen{}\left\{ {#1}_{{}_{}}\,\negthickspace\right\}\mathclose{}}
\newcommand{\abs}[1]{\mathopen{}\left| {#1}_{{}_{}}\,\negthickspace\right|\mathclose{}}
\newcommand{\Var}{\mathbbm{V}\textrm{ar}}
\newcommand{\E}{\mathbbm{E}}

\jmlrheading{1}{2000}{1-48}{4/00}{10/00}{00-000}{Argouarc'h, Desbouvries, Barat, Kawasaki, Dautremer}
\ShortHeadings{Discretely indexed flows}
{Argouarc'h, Desbouvries, Barat, Kawasaki, Dautremer}
\firstpageno{1}

\begin{document}

\title{Discretely Indexed Flows}

\author{
\name Elouan Argouarc'h
\email elouan.argouarc'h@cea.fr\\
\addr Université Paris Saclay, CEA, List, 
\email elouan.argouarch@telecom-sudparis.eu \\
F-91120 Palaiseau, France \\
\addr and \\
\addr Samovar, Telecom SudParis,
Institut Polytechnique de Paris, \\
F-91120 Palaiseau, France \\
\AND
\name Fran\c{c}ois Desbouvries
\email francois.desbouvries@telecom-sudparis.eu \\
\addr Samovar, Telecom SudParis,
Institut Polytechnique de Paris, \\
F-91120 Palaiseau, France \\
\AND
\name Eric Barat
\email eric.barat@cea.fr \\
\name Eiji Kawasaki
\email eiji.kawasaki@cea.fr \\
\name Thomas Dautremer
\email thomas.dautremer@cea.fr\\
\addr Université Paris Saclay, CEA, List, \\
F-91120 Palaiseau, France\\
}

\editor{Elouan Argouarc'h}

\maketitle

\begin{abstract}
In this paper we propose Discretely Indexed flows (DIF)
as a new tool for solving variational estimation problems.
Roughly speaking,
DIF are built as an extension of Normalizing Flows (NF),
in which the deterministic transport 
becomes stochastic, and more precisely discretely indexed.
Due to the discrete nature of the underlying additional latent variable,
DIF inherit the good computational behavior of NF:
they benefit from both a tractable density as well as a straightforward sampling scheme,
and can thus be used for the dual problems of 
Variational Inference (VI) 
and of Variational density estimation (VDE). 
On the other hand,
DIF can also be understood as an extension of mixture density models,
in which the constant mixture weights are replaced by flexible functions.
As a consequence, 
DIF are better suited for capturing distributions with discontinuities, sharp edges and fine details, which is a main advantage of this construction.
Finally we propose a methodology for constructiong DIF in practice, and see that DIF can be sequentially cascaded, and cascaded with NF. 
\end{abstract}


\section*{Introduction}
Many scientific tasks take interest in decision making with respect to (wrt) some random process. In this context, evaluating the probability density function (pdf) and/or obtaining random samples from the process can help the decision making by computing statistical quantities of interest. For example computing confidence intervals may help to conclude on the existence or absence of some underlying effect. Historical methods include posterior inference with MCMC \citep{BayesianComputationMCMC} \citep{UnderstandingMetropolisHastings}
or Approximate Bayesian Computation in the likelihood-free setting \citep{ApproximatedBayesianComputation}.

The task of probabilistic modelling provides with a concurrent approach: by using an approximating distribution (sometimes referred to as a \emph{surrogate}) with either or both a tractable density and an explicit sampling mechanism, we can estimate relevant statistics. This includes the non-parametric approach of Kernel Density Estimation \citep{KDE}. 
Variational probabilistic modeling consists in building a surrogate probability distribution by solving an optimization problem among some parametric family of distributions \citep{StochasticVariationalInference} \citep{ReviewVI} \citep{EMAlgorithm}. 
Recent advances in automatic differentiation \citep{AutomaticDifferentiation} \citep{Pytorch} \citep{Tensorflow} and optimization \citep{ADAM} have paved the way to using Neural Networks (NN) functions in probabilistic modeling \citep{VAE} \citep{GAN} \citep{DDPM}. 
Note however that concurrent approaches can perform density estimation \citep{NeuralDensityRatio} with leveraging neural network functions which approximate a density ratio but without explicitly constructing a probability distribution. 

Normalizing Flows (NF) \citep{reviewNF} \citep{NormalizingFlowsForProbabilisticModeling} are a versatile tool for probabilistic modelling as they allow for both generation of random samples with an explicit sampling mechanism, and density estimation with exact pdf evaluation. Therefore NF are at the crossroad between Variational Inference (VI) \citep{VIwithNF} \citep{IAF}, Variational Density Estimation (VDE) \citep{RealNVP} \citep{MAF} and Generative Modelling \citep{Glow}. 
These three problems are especially relevant in the field of machine learning which explain the popularity of NF amongst machine learning practitionners. Moreover, part of their attractiveness results from the fact that NF define deterministic invertible transformations which can effortlessly be layered to produce deep and flexible families of surrogates, making them competitive on a performance standpoint. 

In this paper we build DIF as an extension of NF, 
and we therefore provide with another method in order to build surrogate probability distributions. 
DIF no longer rely on a deterministic mapping but rather leverage a stochastic transformation,
all the while remaining in the same sweet spot as NF: they allow for both exact pdf evaluation and straightforward sampling.  
On the other hand, DIF can also be seen as an extension of mixture density models, in which the constant mixture weights become flexible functions.
As a result, DIF enable to capture distributions with finer details than regular mixture models. 

The rest of the paper is organised as follows. 
In section \ref{dual}
we present the two dual problems of VI, 
on the one hand, 
and VDE, on the other hand. 
Both are probabilistic modeling problems, in which we build a surrogate $\Psi$
of the true probability distribution $P$; 
in the first case,
we use the pdf $p$ associated to $P$,
and in the second case observed samples from $P$.
In section \ref{NF}, we recall the principles of NF,
explain how they can be used for 
VI and VDE,
and revisit them as latent variables models.

In section \ref{dif} we extend NF to DIF;
roughly speaking, 
the deterministic transport is replaced by a (discrete) stochastic one,
therefore 
DIF are latent variable models too,
but the original latent space is augmented by an additional discrete variable.
From a computational point of view, DIF retain the good behaviour of NF; indeed,
the discrete nature of the additional variable enables for explicit density evaluation as well as a closed form formula
for the reverse transition kernel between the latent and observed spaces. 
We next see that similarly to NF, 
DIF can be used efficiently either for the VI or for the VDE problems;
as far as VI is concerned, our work builds upon the previous Transport Monte Carlo (TMC) approach \citep{TransportMonteCarlo},
but we argue in favor of a more coherent optimization objective than that used in the TMC approach. 

Finally in section \ref{dif_in_practice}
we propose a methodology for constructing DIF in practice. Namely we propose a convenient parameterization of the DIF stochastic transport. 
Under this parameterization, DIF can be considered as an extension of a GMM, 
the benefits of which are illustrated via simulations on complex two-dimensional distributions. 
We finally see that DIF can be combined together (or with NF as well), 
and that they can be used for conditional density estimation.
We end the paper with a conclusion. The full code is available at \url{https://github.com/ElouanARGOUARCH/Discretely-Indexed-Flows}.

\section{Two dual probabilistic modeling problems}

In this section we propose a parallel discussion of the VI and VDE problems, 
which are the two modelling problems addressed in this paper. 

\label{dual}
\subsection{Variational Inference}\label{VI}
Suppose that we dispose of $p(x)$, in a possibly unormalized form, but we do not have a simple procedure for sampling from the distribution $P$. This is usually the case when considering a posterior distribution, the pdf of which is proportional to the product of the \emph{prior} and of the \emph{likelihood},
but
the normalizing constant (the \emph{evidence}) is unavailable. 
VI aims at providing samples that are approximately distributed according to $P$, by considering a variational distribution defined as: 
\begin{equation*}
    \Psi^{\star} = \underset{\Psi}{\arg\min}\, \mathrm{D}^{(\mathrm{VI})}\paren{\Psi,P},
\end{equation*}
where $\mathrm{D}^{(\mathrm{VI})}$ is some discrepancy measure and $\Psi$ belongs to some family of distributions which is straightforward to sample from. 
Since $\Psi^{\star}$
is close to $P$,
samples from $\Psi^{\star}$
are approximately distributed according to P.

Note moreover that if the pdf $\psi^{\star}$ is available, one can use $\Psi^{\star}$ as an importance distribution for targeting $P$.
Furthermore, one can use Rubin's SIR mechanism
\citep{SequentialImportanceResampling}
\citep{gelfand-smith}
\citep{smith-gelfand}
\citep[\S 9.2]{Cappeetal}
to produce asymptotically  independent and identically distributed (i.i.d.) samples from $P$. 

In this paper 
$\mathrm{D}^{(\mathrm{VI})}$ will be a Kullback-Leibler Divergence \citep{KullbackLeibler} ($\mathrm{D}_{\mathrm{KL}}$),
be it either the forward one $\mathrm{D}_{\mathrm{KL}}(P||\Psi)$ or the reverse one $\mathrm{D}_{\mathrm{KL}}(\Psi||P)$. 
We also consider a parametric family $\set{\Psi_\theta|\theta \in \Theta}$. 
However for arbitrary $P$, 
neither 
$\mathrm{D}_{\mathrm{KL}}(P||\Psi)$
nor $\mathrm{D}_{\mathrm{KL}}(\Psi||P)$
admits a closed form expression, which calls for a Monte Carlo (MC) approximation.
Since we can only sample from $\Psi$, the discrepancy measure $\mathrm{D}^{(\mathrm{VI})}$ must be the reverse $\mathrm{D}_{\mathrm{KL}}$, and an MC approximation can be computed as: 
\begin{equation}
\label{DKL_VI}
    \mathrm{D}_{\mathrm{KL}}(\Psi||P) = \E_{\Psi}\paren{\log\paren{\frac{\psi\paren{X}}{p\paren{X}}}}\approx \frac{1}{M}\sum_{\stackrel{i=1}{x_i\sim \Psi}}^{M} \log\paren{\frac{\psi\paren{x_i}}{p\paren{x_i}}}.
\end{equation}
Minimizing this MC estimate wrt model parameters $\theta$ via Gradient Descent (GD) requires that $p$ is differentiable (which is assumed throughout this paper), and also that $\psi$ is chosen to be differentiable wrt $\theta$.
However computing gradients can still be challenging
because the samples $x_i\sim \Psi$ indeed depend on model parameter $\theta$. 
One way to compute the gradients is to use a \emph{reparameterization trick},
that is,
to use an invertible differentiable standardization function $S(.;\theta)$ such that random variable (rv) $S(X;\theta) = \epsilon$, does not depend on $\theta$. 
Then re-writing $x_i$ as  $x_i = S^{-1}\paren{\epsilon_i; \theta}$
enables to compute the gradients of \eqref{DKL_VI} wrt $\theta$.

\subsection{Variational Density Estimation}\label{DE}

Suppose  that we dispose of samples $x_1,..., x_M \sim P$ but we cannot evaluate the pdf $p(x)$.
This occurs for example when we have only recorded observations from an otherwise unknown real-world stochastic process. 
Among other techniques,
we can perform VDE to obtain an estimation of $p(x)$ by considering a variational distribution defined as: 
\begin{equation*}
    \Psi^{\star}= \underset{\Psi}{\arg\min}\, \mathrm{D}^{(\mathrm{VDE})}\paren{\Psi,P},
\end{equation*}
where $\mathrm{D}^{(\mathrm{VDE})}$ is some discrepancy measure and $\Psi$ belongs to some family of distributions with tractable pdf. 
Since $\Psi^{\star}$
is close to $P$, pdf $\psi^{\star}$ is an estimate of the unknown density function $p$. 

Note moreover that if $\Psi^{\star}$ is easy to sample from, then samples from $\Psi^{\star}$ are approximately distributed according to $P$.

Once again, we will consider $\mathrm{D}^{(\mathrm{VDE})}$ to be a $\mathrm{D}_{\mathrm{KL}}$  and the parametric family $\set{\Psi_\theta|\theta \in \Theta}$. 
Minimizing 
an MC approximation of the reverse $\mathrm{D}_{\mathrm{KL}}$,
as in \eqref{DKL_VI},
is not possible here. Indeed in this case,
pdf $p$ is evaluated at samples points $x_i \sim \Psi$, which depend on $\theta$; hence we need to account for the terms $p(x_i)$ in the optimization, which is not possible since function $p(.)$ is unknown.
This calls for the use of the \emph{forward} $D_{\mathrm{KL}}$, and an MC approximation using the samples from $P$ can be computed as : 
\begin{equation*}
    \mathrm{D}_{\mathrm{KL}}(P||\Psi) = \E_{P}\paren{\log\paren{\frac{p\paren{X}}{\psi\paren{X}}}}\approx \frac{1}{M}\sum_{\stackrel{i=1}{x_i\sim P}}^{M} \log\paren{\frac{p\paren{x_i}}{\psi\paren{x_i}}}.
\end{equation*}
Though this MC estimate of the $\mathrm{D}_{\mathrm{KL}}$ cannot be computed since $p$ is not available, 
note that $x_i$ and $p(x_i)$ do not depend on $\theta$,
and can thus be ignored in the minimization process.
As a consequence minimizing this MC approximation of the $D_{\mathrm{KL}}$ reduces to maximizing the log-likelihood of the data under model $\Psi$. Finally, minimizing the MC approximation of $\mathrm{D}_{\mathrm{KL}}(P||\Psi)$ can be conducted via GD, which only requires that $\psi$ is differentiable
(here, unlike in section \ref{VI} the samples do not depend on $\theta$, so the gradients can be computed directly).

\section{Normalizing Flows}\label{NF}

In this section we propose a brief presentation of NF, which have been first introduced in \citep{VIwithNF}
(see also \citep{reviewNF} \citep{NormalizingFlowsForProbabilisticModeling} for thorough reviews of the topic). We explain how to use NF for the two problems of VI and VDE.

\subsection{Change of variables, sampling mechanism and density evaluation}
\label{change variables}

The underlying idea of NF is that of a bijective change of variables. 
Let $U$ and $V$ be two rv related via:
\begin{equation}
\label{ChangeOfVariable}
    V = f^{-1}\paren{U}
\end{equation}
for some  
C1-diffeomorphism $f$, 
that is,
an invertible mapping such that both $f$ and its inverse $f^{-1}$ are differentiable and with continuous derivatives. 
Let $q_U$ and $q_V$ be respectively the pdf of $U$ and $V$.
As is well known, \eqref{ChangeOfVariable} induces:
\begin{equation}
\label{ChangeOfVariableDensity}
    q_V(x) = q_U\paren{f(x)}\abs{\det \mathrm{J}_f(x)},
\end{equation}
where $\mathrm{J}_f$ is the Jacobian matrix.
This change of variables formula for densities in fact defines the pdf $q_V$ via a functional transform $\mathcal{F}$ of $q_U$ and of the mapping $f$: 
\begin{equation}
\label{functionalFlow}
    q_V = \mathcal{F}\paren{q_U;f}
\end{equation}

These formulas are potentially useful for sampling \eqref{ChangeOfVariable} and for density evaluation \eqref{ChangeOfVariableDensity}.
However, at this point it is interesting to observe that they do not involve the same assumptions on $q_U$ and $f$:
\begin{enumerate}
    \item 
    If $f^{-1}$ is available without tears, and if $q_U$ is easy to sample from,
    then \eqref{ChangeOfVariable} can be used as a straightforward sampling mechanism: if $u\sim q_U$ then $v= f^{-1}\paren{u}\sim q_V$, in other words we first sample $u$ from $q_U$ and then map $u$ to $v$ via $f^{-1}$;
    \item
    On the other hand, evaluating $q_V$ via \eqref{ChangeOfVariableDensity} requires that pdf $q_U$ can be evaluated at any point and that we can compute the Jacobian determinant easily. 
\end{enumerate}
\subsection{Application to the variational problems}
\label{appli12}

Let us now see how to apply (\ref{ChangeOfVariable}) and (\ref{ChangeOfVariableDensity}) to the 
variational problems identified in 
sections \ref{VI}
and \ref{DE}. 
Given a $Q$ distributed rv $Z$ (usually chosen as a fixed standard Gaussian distribution $\mathcal{N}\paren{0,\mathrm{I}_d}$ - as discussed in section \ref{whyNF} - where $d$ is the dimension of the problem), both problems consist in designing a change of variable $T$ such that the distribution of $\widetilde{X} = T^{-1}\paren{Z}$, which we denote as $\Psi$, is close to target $P$ (in the sense of the appropriate $D_{\mathrm{KL}}$). In the general case and for both problems of VDE and VI, the optimization problem will not have a solution for arbitrary $T$. 
Therefore, we consider $T \in \set{T_\theta|\theta\in\Theta}$,
with the condition that $T$ is differentiable wrt model parameters $\theta$,
in order to solve the optimization using GD
(see 
\citep{NICE} 
\citep{RealNVP}
\citep{MAF}
\citep{IAF}
\citep{MAF}
\citep{Glow}
\citep{NeuralSplineFlows}  for examples of such parametrization).

\begin{figure}[h]
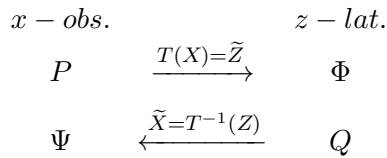

\[
\begin{array}{ccc}
x-obs. && z-lat. \\
P& 
\xrightarrow{T\paren{X} = \widetilde{Z}}&
\Phi
\\[0.2cm]
\Psi&
\xleftarrow{\widetilde{X} = T^{-1}\paren{Z}}&Q
\end{array}
\]
\caption{Forward \& Backward mappings between observed and latent spaces}
\label{fig:4-lois-NF}
\end{figure}

The fact that mapping $T$ is a C1-diffeomorphism
has interesting consequences.
First, mapping $T$ indeed provides two couples of rv:
$\bracket{\widetilde{X} = T^{-1}\paren{Z},Z}$
(see the second row of figure \ref{fig:4-lois-NF}), but also
$\bracket{X, \widetilde{Z} = T\paren{X}}$
(see first row),
in which $X \sim P$ and $\Phi$ denotes the distribution of 
$\widetilde{Z}$.
Then, if $Q$ and $P$ respectively admit pdf $q$ and $p$, the pdf $\psi$ and $\phi$ associated with $\Psi$ and $\Phi$ are defined via the same functional transform \eqref{functionalFlow}:
\begin{align}
    \label{fonctionnelle-1-psi}
    \psi & 
    =\mathcal{F}\paren{q; T} 
    \\
    \label{fonctionnelle-1-phi}
    \phi & 
    =\mathcal{F}\paren{p; T^{-1}}
\end{align}

Moreover, by applying the simple change of variable $z = T(x)$, we have the following two equalities: 
\begin{eqnarray}
   \label{DKLEqualityReverseObserved}
   D_{\mathrm{KL}}\paren{\Psi||P} &=& D_{\mathrm{KL}}\paren{Q||\Phi},
\\
\label{DKLEqualityForwardObserved}
    D_{\mathrm{KL}}\paren{P||\Psi} &=& D_{\mathrm{KL}}\paren{\Phi||Q}. 
     \end{eqnarray}
These two equalities explain that, 
since observed and latent distributions are related via a deterministic invertible mapping, 
minimizing a forward (resp. reverse) $D_{\mathrm{KL}}$ in the latent space (that of $\widetilde{Z}$ and $Z$) mechanically minimizes a reverse (resp. forward) $D_{\mathrm{KL}}$ in the observed space (that of $X$ and $\widetilde{X}$), and vice-versa.
These remarks will be useful in later sections.

\subsubsection{Why {\em Normalizing}? Why {\em Flow}?}
\label{whyNF}
\begin{itemize}
\item
With an argument similar to the inverse cumulative distribution function (CDF) technique for sampling \citep[\S 2.2]{NormalizingFlowsForProbabilisticModeling}, for any distribution $P$ with compact support, we can (at least theoretically) construct a transport $T$ between $P$ and a standard Normal distribution (whence the term \emph{Normalizing} Flows). For that reason, it is routinely assumed to set $Q$ as a standard parameter-free Normal distribution (which is assumed throughout the rest of this paper); and using an NF for modeling P reduces to approximating a combination of two CDF. 

\item
 In order to ensure sufficient flexibility in $T$, we leverage the property that C1-diffeomorphisms are closed under composition. If we define for example $T = T_1 \circ T_2$, where $T_1, T_2$ are C1-diffeomorphisms, then $T$ is also a C1-diffeomorphism and its Jacobian determinant can be computed using the chain rule formula: 
\begin{align*}
    |\det \mathrm{J}_T(x)| & = |\det \mathrm{J}_{T_1}(x)| \times |\det \mathrm{J}_{T_2}\paren{T_1(x)}|,
 \end{align*}
 which implies $\mathcal{F}\paren{.;T} = \mathcal{F}\paren{\mathcal{F}\paren{.; T_2}; T_1}$. Hence, it is easy to construct $T$ as a composition of simple transformations $\{T_c\}_{c=1,...,C}$. 
 The distribution $Q$ sequentially gets morphed into $\Psi$ by a \emph{Flow} of transformations, whence the term Normalizing {\em Flows}.
 \end{itemize}

\subsubsection{VI with NF}
\label{appli1}

\begin{figure}[h]
\caption{Example of VI using a multi-step NF}
\label{fig:FlowVI}
\centering
\includegraphics[width=\textwidth]{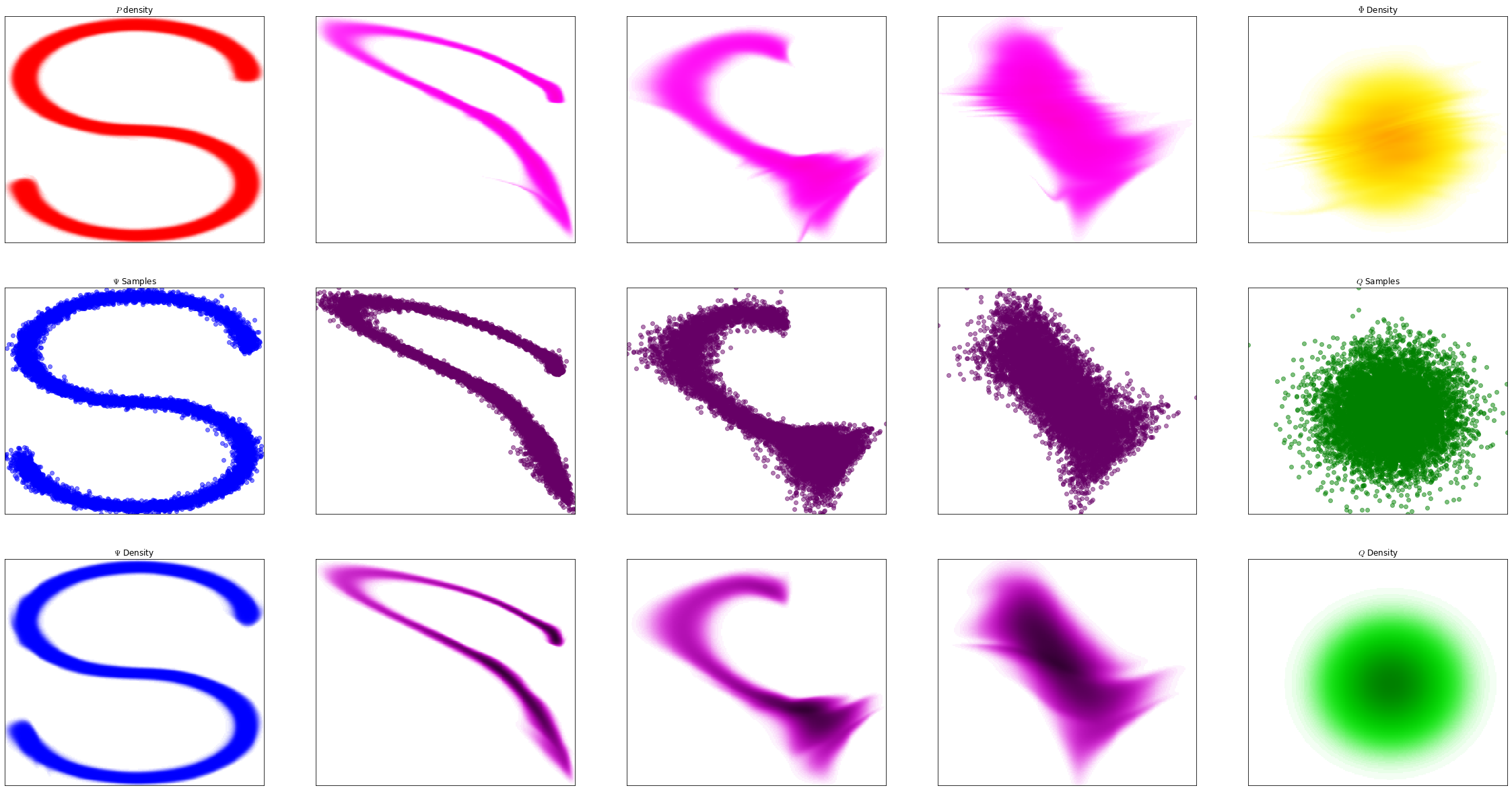}
\end{figure}

Let us consider the VI setting described in section \ref{VI}. Minimizing an MC approximation of the reverse $D_{\mathrm{KL}}$ leads to the following optimization problem: 
\begin{equation}
\label{maxi-nf-vi}
    \min_{\theta \in \Theta} \sum_{\stackrel{i=1}{x_i\sim \Psi}}^{M}\log\paren{\frac{\psi(x_i)}{p(x_i)}}.
\end{equation}
In order for this objective to be differentiated wrt $\theta$ we may try to apply a reparametrization trick.
Equation \eqref{DKLEqualityReverseObserved} hints at such a reparameterization: since $x_i = T^{-1}(z_i)$, where $z_i \sim Q$ does not depend on $\theta$ (because we have assumed that $Q$ is parameter-free) and $T$ is differentiable with respect to model parameters $\theta$. Hence by construction, NF provides with a  straightforward differentiable reparametrization trick $x = T^{-1}(z)$. By applying this change of variable, \eqref{maxi-nf-vi} becomes:
\begin{align}
    \nonumber
    \min_{\theta\in\Theta}& \sum_{\stackrel{i=1}{z_i\sim Q}}^{M}\log\paren{\frac{\psi\paren{T^{-1}(z_i)}}{p\paren{T^{-1}(z_i)}}} = \min_{\theta\in\Theta} \sum_{\stackrel{i=1}{z_i\sim Q}}^{M}\log\paren{\frac{q(z_i)}{\phi(z_i)}}  \\
    \label{maxi-pb1}
    &  =\max_{\theta\in\Theta} \sum_{\stackrel{i=1}{z_i\sim Q}}^{M}\log\paren{\phi\paren{z_i}} \stackrel{\eqref{fonctionnelle-1-phi}}{=} \max_{\theta\in\Theta} \sum_{\stackrel{i=1}{z_i\sim Q}}^{M}\log\paren{p\paren{T^{-1}\paren{z_i}}\abs{\det \mathrm{J}_{T^{-1}}(z_i)}}.
\end{align}
The resulting optimization problem can be solved through Gradient Ascent (GA) as this expression is differentiable with respect to $\theta$, and $Q$ was purposely chosen to be easy to sample from. 
Let $\theta^{\star}$ maximize \eqref{maxi-pb1};
it remains to sample the corresponding model $\Psi^*$ to produce samples that are approximately distributed according to $P$.

Figure \ref{fig:FlowVI} presents an example of an NF used for VI on a 2-dimensional S-Curve problem. The left most column shows the observed-$x$ space while the right most column corresponds to the latent-$z$ space. The model is defined as a composition of 4 Real NVP coupling layers \citep{RealNVP}, the middle columns present the intermediate distributions between each transformation. We can therefore visualise how a standard Gaussian distribution $Q$ (green) is sequentially morphed into the model distribution $\Psi$ (blue) that resembles the target $P$ (red). As expected, $\Phi$ (yellow) resembles $Q$. The first row shows a color-mapping of the target density $p$ function getting morphed via $T$. The last two rows are both representations of $\Psi$ but the first shows drawn samples while the later is a color-mapping of the density function $\psi$ which is a result of $q$ getting morphed via $T^{-1}$. 

\subsubsection{DE with NF}
\label{appli23}

\begin{figure}[h]
\caption{Example of Density Estimation and Sampling using a multi-step NF}
\label{fig:FlowDE}
\centering
\includegraphics[width=\textwidth]{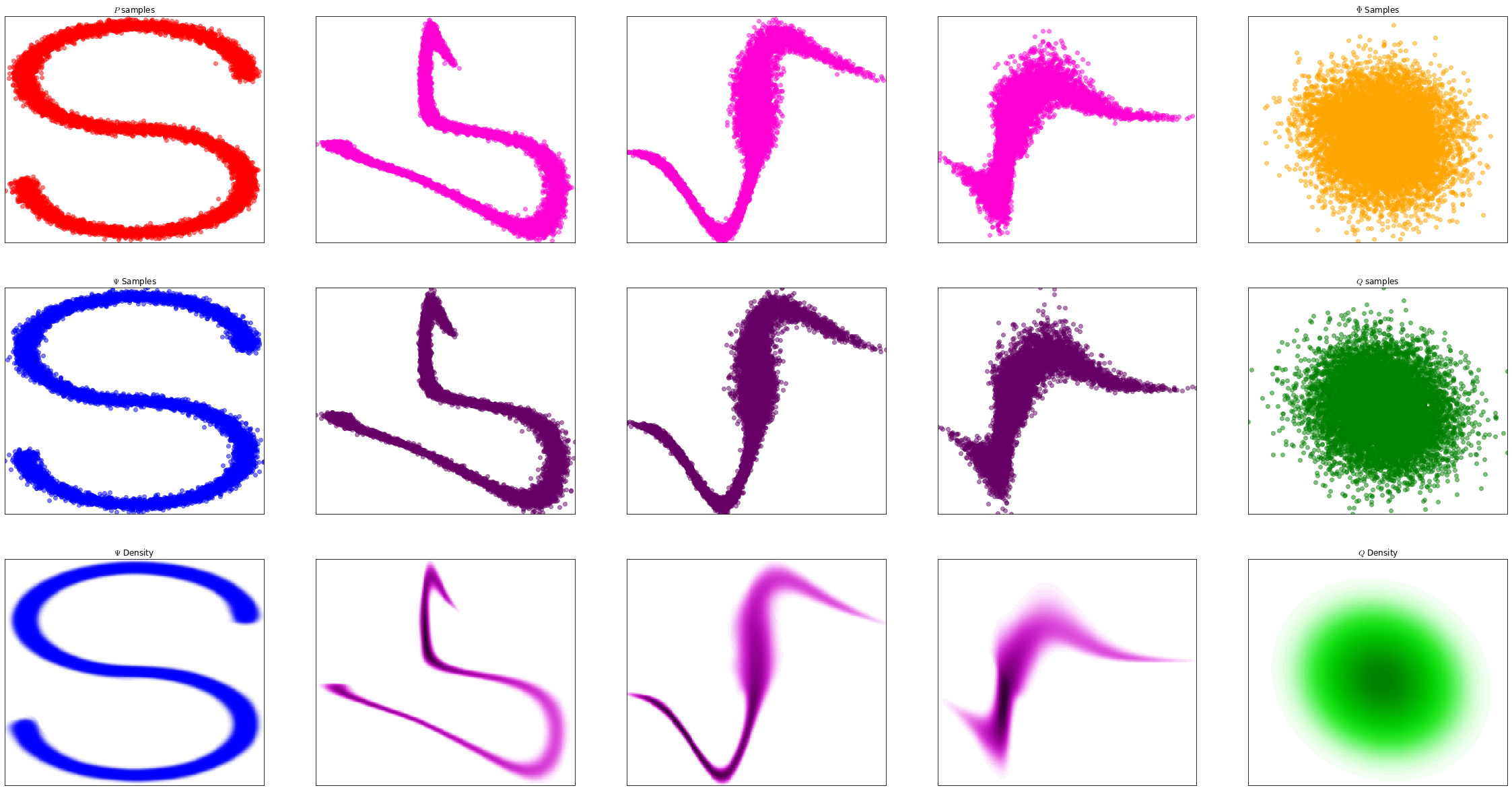}
\end{figure}

Consider now the VDE setting described in section \ref{DE}, 
the Maximum-Likelihood Estimation (MLE) problem (which we recall is equivalent to minimizing an MC approximation of the forward $D_{\mathrm{KL}}$) reads:
\begin{equation}
\label{MaxLogLikelihood}
    \max_{\theta\in\Theta} \sum_{\stackrel{i=1}{x_i\sim P}}^{M} \log\paren{\psi\paren{x_i}} \stackrel{\eqref{fonctionnelle-1-psi}}{=}\max_{\theta\in\Theta} \sum_{\stackrel{i=1}{x_i\sim P}}^{M} \paren{q\paren{T(x)}\abs{\det \mathrm{J}_T(x)}}
\end{equation}
For this optimization objective to be differentiable, it is only necessary that the density function $q$ is also differentiable, which is the case since $Q$ is a standard Normal distribution. Then, the maximization can be solved through GA.
Let $\theta^{\star}$ maximize equation \eqref{MaxLogLikelihood};
then model pdf $\psi^*(x)$ is an approximation of the target pdf $p$ and hence solves the VDE problem. 
Note moreover that we can produce new samples that are approximately distributed from $P$ by sampling the corresponding model $\Psi^*$.

Figure \ref{fig:FlowDE} presents an example of an NF used for VDE on the same target distribution as in figure \ref{fig:FlowVI}. The flow model is also defined as a composition of 4 Real NVP layers. The only difference with Figure \ref{fig:FlowVI} is that the target distribution is available via its samples and therefore the first row shows the samples of $P$ being transformed via $T$, the interpretation of this figure is otherwise the same.

\subsection{Topological limitations}
\label{topological}
\begin{figure}[!h]
    \includegraphics[width=\textwidth]{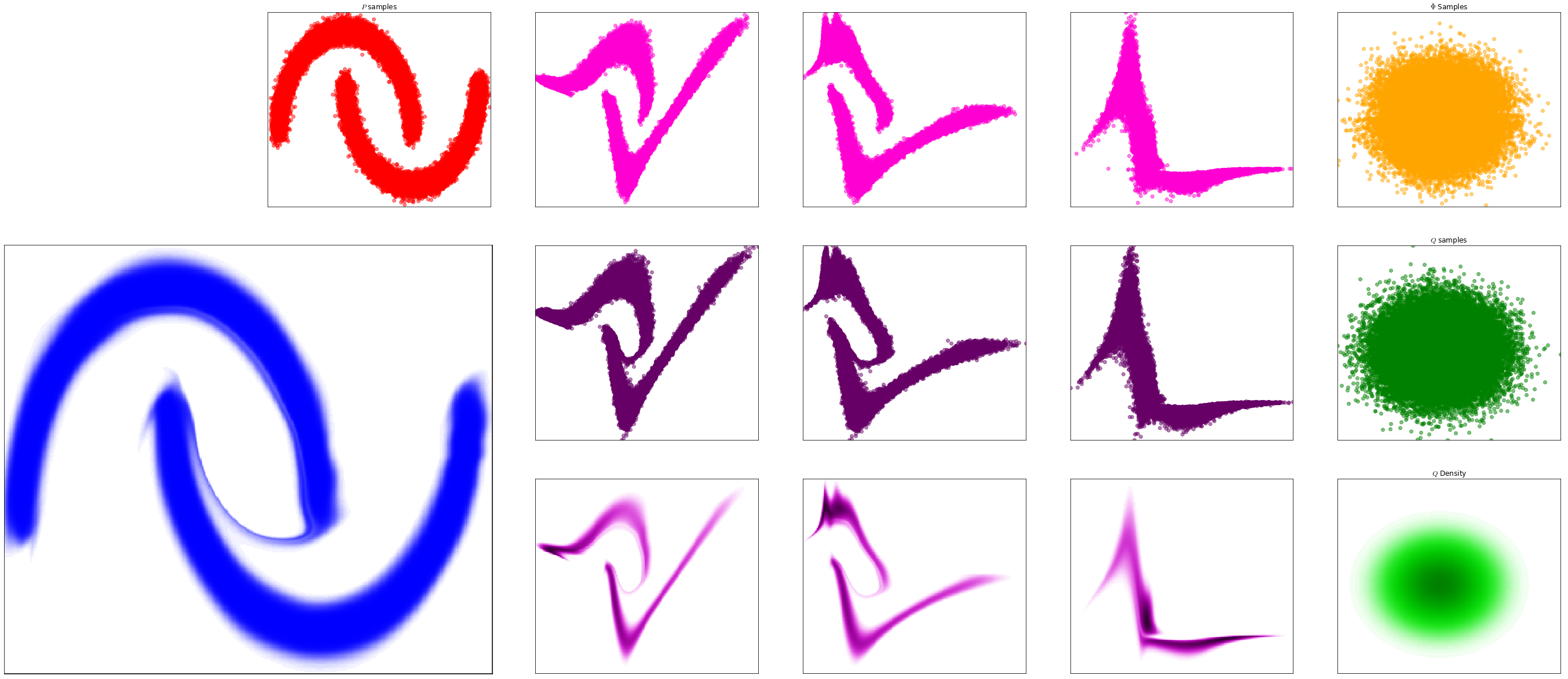}
\caption{Topological limitation of an NF; it struggles to approach distribution with disjoint support}\label{fig:constraint}
\end{figure}

Observe however that, by essence, NF are not well suited to approximate multimodal distributions with disjoints supports: since $T^{-1}$ continuously reshapes the Normal distribution $Q$  into $\Psi$, the model will struggle to efficiently disband the mass into several modes. We illustrate this in figure \ref{fig:constraint} where we try using a multi-step NF to approach a distribution with two disjoint moon elements. In this context, we see that there remains an artefact connection between the two elements of mass; hence the resulting NF distribution is not one with disjoint supports. This topological limitation is one drawback of NF models which DIF circumvent, see section \ref{breakbarrier} below. 
 \newline

\subsection{Towards Discretely Indexed Flows}

As we now see,
NF can be considered as latent variable models,
which suggests the extension to DIF
which will be addressed in section 
\ref{dif}.

\subsubsection{Flows as latent variable models}

NF target a distribution $P$ by constructing a distribution $\Psi$ associated with a rv $\widetilde{X} = T^{-1}\paren{Z}$. The rv $Z$ is a proxy latent variable distributed according to $Q$, and we adjust $T$ so that $\Psi$ is as close as possible of $P$. $\Psi$ is therefore the marginal distribution of interest, out of a couple of rv $(\widetilde{X},Z)$ with joint density $\overleftarrow{\pi}\paren{x|z}q(z)$. 
It can thus be considered as a latent variable model:
we are given a prior $q$ (the distribution of the latent variable $Z$), and we move from $z$ to $x$ via the conditional distribution $x\sim\overleftarrow{\Pi}\paren{.|z}$. 
Of course, since $Z$ and $\widetilde{X}=T^{-1}\paren{Z}$ are related via a {\sl deterministic} mapping, the associated conditional density function $\overleftarrow{\pi}\paren{x|z}$ reads:
\begin{equation}
\label{deltaFlow}
    \overleftarrow{\pi}\paren{x|z} = \delta_{T^{-1}\paren{z}}(x).
\end{equation}

\subsubsection{Beyond NF}

One way of increasing expressiveness is to consider a latent variable model in which the deterministic mapping \eqref{deltaFlow} used in NF is replaced by a stochastic transport described by a transition kernel $\overleftarrow{\Pi}$, be it a pdf or a probability mass function. 
In either case, due to the latent variable structure, sampling from $\Psi$ is (almost) as easy as in the deterministic case: we start off by sampling the latent distribution $z \sim Q$, and next we sample from the conditional distribution $x \sim \overleftarrow{\Pi}\paren{.|z}$. Therefore, as long as the prior and likelihood
are both easy to sample from, 
the model can be sampled from effortlessly. \newline

The density associated with
$\Psi$ is given by: 
\begin{equation*}
    \psi(x) = \E_{Z\sim Q} \bracket{\overleftarrow{\pi}\paren{x|Z}}  
\end{equation*}

This density is not necessarily tractable as the expectation does not always admit a closed form expression, at least if the conditional distribution is continuous. If however $\overleftarrow{\pi}\paren{x|z}$ is discrete and has finite support, then the integral becomes a tractable sum. \newline

As a consequence, note that for continuous latent variable models, we might not be able to use explicit evaluation of the density function $\psi$ in order to optimize the model. For example, if $\psi$ is untractable, we cannot perform direct MLE of model parameters and we have to rely on more sophisticated optimization procedures. For instance \citep{VAE} presents the Variational AutoEncoder (VAE), which is a continuous latent variable model that leverages a Variational Expectation-Maximization optimization scheme.\newline

However, for the task of VDE, it is desirable that we use a model $\Psi$ s.t. pdf  $\psi$ exists and can be computed exactly. Therefore, in the rest of the paper, we will explore DIF which is a class of latent variable models that builds upon the principle of invertible mappings used in NF and includes stochasticity in a discrete form to ensure tractable density.

\section{DIF}
\label{dif}

In this section we introduce DIF 
as one possible 
stochastic extension of NF. 
More precisely, we build DIF as a latent variable model where the deterministic mapping between $z$ and $x$ is replaced by a discrete stochastic distribution. 
We then apply DIF for both the VI and VDE problems.

\subsection{DIF as a discrete latent variable model}\label{DIF_as_LVM}

We define a DIF model via some prior distribution $Q$ and the likelihood:
\begin{equation}\label{DIF_likelihood}
    \overleftarrow{\pi}\paren{x|z} = \sum_{k=1}^K w_k(z) \delta_{T_k^{-1}(z)}(x),
\end{equation}
where $\{w_k, T_k\}_{k=1,...,K}$ are specified parameterized functions. With words: for a given value of $z$, $z$ is transformed into $x = T_k^{-1}(z)$ via mapping $T_k$ with probability $w_k(z)$. The function $w_k(z)$ therefore represents the conditional probability $\mathbbm{P}\paren{x = T_k^{-1}(z)|z}$ and must sum to 1: $\sum_{k=1}^K w_k(z) = 1, \forall z\in\mathbbm{R}^d$.\newline

DIF in fact is an auxiliary latent variable model where we leverage a categorical latent variable $U$ to create a stochastic transport instead of a deterministic one. The latent rv $U$ takes discrete values 1 to $K$ which indicate what mapping is applied to $z$. The resulting rv $\widetilde{X}$ can be written as :
\begin{equation}
\label{DIF-chgt-de-variable}
\widetilde{X} = T_U^{-1}(Z) \text{ where } Z\sim Q \text{ and } U \sim {\rm{Categorical}}\paren{w_1(Z),..., w_K(Z)}.
\end{equation} 
Hence, sampling from this model remains straightforward as the stochastic transport of prior samples $z \sim Q$ can be conducted with sampling $\overleftarrow{\Pi}(.|Z=z)$ using \eqref{DIF-chgt-de-variable}. DIF is therefore a viable parametric model candidate for VI (see section \ref{DIFVI} below).\
\begin{center}
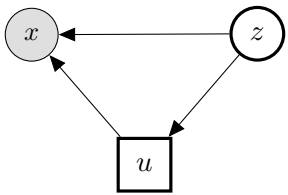

    \begin{tikzpicture}[transform shape, node distance=1cm, roundnode/.style={circle, draw=black, very thick, minimum size=7mm},squarednode/.style={rectangle, draw=black, very thick, minimum size=7mm}][h]
        \node[squarednode](u){$u$};%
        \node[obs](x)[above= of u, xshift = -1.5cm]{$x$};%
        \node[roundnode][above=of u,xshift=1.5cm](z){$z$}; %
        \edge {u,z}{x}  
        \edge{z}{u}
    \end{tikzpicture}
    \captionof{figure}{DIF Graphical Model - Backward transition $\protect\overleftarrow{\Pi}$}
\end{center}

With this choice for $\overleftarrow{\Pi}$ and with the additional constraint that each $T_k$ for $k=1,...,K$ is a C1-diffeomorphism, 
one can show (see Appendix \ref{Psi-and-reverse-kernel}) 
that the marginal pdf $\psi$ reads
\begin{equation}\label{DIFDensity}
   \psi(x) = \sum_{k=1}^Kw_k\paren{T_k(x)}q\paren{T_k(x)}\abs{\det \mathrm{J}_{T_k}(x)}.
\end{equation}Once again, this pdf is therefore defined as a functional transform of the prior pdf $q$ and of functions $\{w_k, T_k\}_{k=1,...,K}$, which
we denote similarly as
\begin{equation}
\label{fonctionnelle-DIF}
    \psi = \mathcal{F}\paren{q; \overleftarrow{\Pi}}. 
\end{equation}
Since \eqref{DIFDensity} can be computed in closed-form, 
DIF can be used to tackle VDE
(see section \ref{DIFDE} below).

At this point, let us observe that DIF can be seen as an extension of two different classes of models:
\begin{itemize}
    \item 
For $K=1$, 
the stochastic transform becomes deterministic,
and indeed the DIF reduces to an NF;
\item
A DIF with $K>1$ components but with constant functions $w_k$
is nothing but a mixture model,
since in this case
the categorical latent variable $U$ does not depend on $z$. This point of view will prove of particular interest in section \ref{DIFvsMixture}.
\end{itemize}

\subsection{Back and forth between observed and latent space}

Recall that for NF the transport was deterministic and invertible, so we were able to go back and forth between observed and latent spaces by applying either $T$ or $T^{-1}$ (see section \ref{appli12}).

In the case of DIF, 
for a given value $z$, 
$x$ is one of the values $\set{T_1^{-1}(z),...,T_K^{-1}(z)}$ 
with associated probabilities $\set{w_1(z),...,w_K(z)}$;
and similarly,
for a given $x$, 
$z$ must be one of the values $\set{T_1(x),...,T_K(x)}$ 
with associated probabilities $\set{v_1(x),...,v_K(x)}$.
Indeed one can show (see Appendix \ref{Psi-and-reverse-kernel}) that the forward transport $\overrightarrow{\Pi}$ reads:
\begin{eqnarray}
    \nonumber
    \overrightarrow{\pi}\paren{z|x}  
    &=& 
    \sum_{k=1}^K v_k(x)\delta_{T_k(x)}(z) , \\
    \label{v_k}
    v_k(x)
    &= 
    &\frac{w_k\paren{T_k(x)}q\paren{T_k(x)}\abs{\det \mathrm{J}_{T_k}(x)}}{\sum_{j=1}^Kw_j\paren{T_j(x)}q\paren{T_j(x)}\abs{\det \mathrm{J}_{T_j}(x)}}.
\end{eqnarray}
We now define the rv $\widetilde{Z}$ (and denote $\Phi$ its probability distribution): 
\begin{equation}\label{DIF-reverse-changement-de-variable}
    \widetilde{Z} = T_U(X) \text{ where } X\sim P \text{ and } U\sim {\rm{Categorical}}\paren{v_1(X),...,v_K(X)}.
\end{equation}
So using DIF is almost as convenient as using NF: even though the transportation $\overleftarrow{\Pi}$ from $z$ to $x$ is stochastic, $\overrightarrow{\Pi}$ remains of the same (discrete) nature. The interest of this result is twofold:
\begin{itemize}
\item 
If we dispose of $x \sim P$, we can easily obtain a sample from $\Phi$ by applying \eqref{DIF-reverse-changement-de-variable}.
\begin{center}
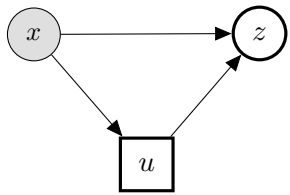

    \begin{tikzpicture}[transform shape, node distance=1cm, roundnode/.style={circle, draw=black, very thick, minimum size=7mm},squarednode/.style={rectangle, draw=black, very thick, minimum size=7mm}][h]
        \node[squarednode](u){$u$};%
        \node[obs](x)[above= of u, xshift = -1.5cm]{$x$};%
        \node[roundnode][above=of u,xshift=1.5cm](z){$z$}; %
        \edge {u,x}{z}  
        \edge{x}{u}
    \end{tikzpicture}
    \captionof{figure}{Forward transition $\protect\overrightarrow{\Pi}$}
\end{center}
Therefore, since we can sample easily from both the likelihood (backward transport $\overleftarrow{\Pi}$) and the posterior (forward transport $\overrightarrow{\Pi}$),
{\sl }we can go back and forth between observed and latent spaces just like in NF. 
The following diagram summarizes the discussion, and should be compared to figure \ref{fig:4-lois-NF}.
\begin{figure}[h!]
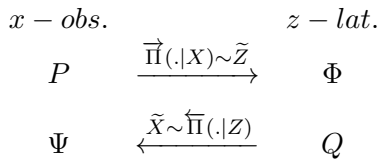

\[
\begin{array}{ccc}
x-obs. && z-lat. \\
P& 
\xrightarrow{\overrightarrow{\Pi}(.|X)\sim\widetilde{Z}}&
\Phi
\\[0.2cm]
\Psi&
\xleftarrow{\widetilde{X}\sim\overleftarrow{\Pi}(.|Z)}&
Q
\end{array}
\]
\caption{Forward \& Backward transitions between obs. and lat. spaces}
\label{fig:4-lois-DIF}
\end{figure}
\item
One can show easily that
\begin{equation}
    \label{fonctionnelle-DIF-reverse}
    \phi = \mathcal{F}\paren{p;  \overrightarrow{\Pi}}.
\end{equation}
Similarly to NF,
both model densities $\psi$ and $\phi$ can thus be written as the result of the same functional transform 
(compare 
\eqref{fonctionnelle-DIF} to
\eqref{fonctionnelle-DIF-reverse},
\eqref{fonctionnelle-DIF} to
\eqref{fonctionnelle-1-psi}
and
\eqref{fonctionnelle-DIF-reverse} to
\eqref{fonctionnelle-1-phi}). Moreover, if we dispose of the pdf $p$ associated with $P$, 
we can compute $\phi$ explicitly.
\end{itemize}
Note that $\mathcal{F}(\phi; \overleftarrow{\Pi}) \ne p$,
which we can relate to the fact that
the backward transportation in the $(X, \widetilde{Z})$ joint distribution is not $\overleftarrow{\Pi}$. 
Instead, for purpose of later arguments, 
let us denote $\overleftarrow{\Pi}'$ the backward transport such that $P = \mathcal{F}(\phi;\overleftarrow{\Pi}')$ and hence: 
\begin{equation*}
    p(x) \overrightarrow{\pi}(z|x) = \overleftarrow{\pi}'(x|z) \phi(z).
\end{equation*}

\subsection{Comparing to the TMC approach}

In this section
we briefly recall some alternate extensions of NF which have been introduced previously.
In section \ref{TMC_approach}
we particularly focus on the TMC approach,
which is closely related to our work;
this section will be of particular interest in section \ref{DIFVI},
where we will further extend the comparison between the two approaches under the scope of the VI problem. 

\subsubsection{Related work}

There have been several prior works which attempted at constructing extensions of NF by using non-deterministic transformations. 

Continuously Indexed Flows (CIF) consider a hierarchical latent variable model with a continuous indexing latent variable. 
CIF have been applied to both VI in \citep{VIwithContinuouslyIndexedFlow} and generative modeling settings \citep{ContinuouslyIndexedFlowGenerative}. However, due to the continuous nature of the augmenting rv, CIF do not admit a tractable density.

Augmented Normalizing Flows (ANF) \citep{AugmentedNormalizingFlows} augment the observation with a continuous rv and use a deterministic NF in order to learn the joint density. 
ANF produce an augmented likelihood which does not allow for exact density evaluation. Both CIF and ANF are classes of models which include the VAE \citep{VAE} and, due to their intractable density, cannot be trained via direct likelihood maximization. Instead, just like VAE, the training consists in maximizing the likelihood via a Variational Expectation-Maximization scheme. 
SurVAE \citep{SurVAE} aims at providing a unified framework for building complex generative models with the use of surjective and stochastic transformations (of which CIF, ANF and DIF are instances). Perhaps more closely related to DIF, \citep{RADapproach} considers a piecewise invertible flow-type transformation which also corresponds to using a discrete indexing variable, 
but where the induced partitioning is hard. This approach allows for a tractable density and does not require summing over the discrete indexing variable since only one of the component is non-negative for any observation. 

\subsubsection{The particular case of TMC}\label{TMC_approach}

In particular, our work can be connected to the previous TMC approach \citep{TransportMonteCarlo}. 
As we shall see in this section, though DIF and TMC use similar stochastic constructions, we will argue in favor of DIF which can be applied to both problems of VI and VDE, while TMC is only suited for the VI setting. Moreover, specifically in a VI setting, the TMC approach considers a particular optimization objective. In section \ref{DIFVI} we will continue the comparison between DIF and TMC in order to discuss the motivation of this optimization objective, and finally in section \ref{RBVI} we will propose a more coherent optimization objective which can be applied to both DIF and TMC. 

The TMC approach is closely related to the methodology proposed by DIF in the sense that it considers a stochastic transport of the same nature as DIF. The definition of the model is however done in a different order as compared to DIF. Indeed, in the TMC approach, the starting point is the target pdf $p(x)$ to which we apply a forward stochastic transport of the form $\overrightarrow{\pi}(z|x) = \sum_{k=1}^K v_k(x)\delta_{T_k(x)}(z)$. We therefore obtain a joint distribution $(X, \widetilde{Z})$ with marginal $\phi(z)$
given by \eqref{fonctionnelle-DIF-reverse}. From this joint distribution, we can computed the associated backward transport:
\begin{eqnarray}
    \nonumber
    \overleftarrow{\pi}\paren{x|z}  
    &=& 
    \sum_{k=1}^K w_k(z)\delta_{T_{k}^{-1}(z)}(x) , \\
    \label{w_k_TMC}
    w_k(z)
    &= 
    &\frac{v_k\paren{T_k^{-1}(z)}p\paren{T_k^{-1}(z)}\abs{\det \mathrm{J}_{T_k^{-1}}(z)}}{\sum_{j=1}^K v_j\paren{T_{j}^{-1}(z)}p\paren{T_{j}^{-1}(z)}\abs{\det \mathrm{J}_{T_{j}^{-1}}(z)}}.
\end{eqnarray}
Finally, the model distribution in the observed space is defined as $\Psi$ which corresponds to $Q$ transported via the the backward transport $\overleftarrow{\Pi}$. Note again that the forward transportation in $(\widetilde{X},Z)$ is not $\overrightarrow{\Pi}$ and we instead denote $\overrightarrow{\Pi}'$. 
\\
\begin{figure}[h!]
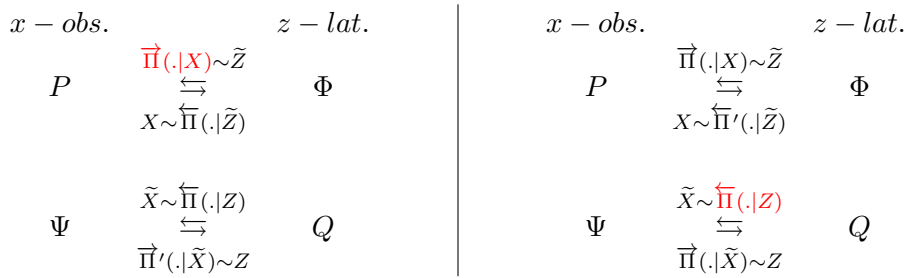

\centering
\begin{subfigure}[b]{0.45\textwidth}
\[
\begin{array}{ccc}
x-obs. && z-lat. \\
P &\overset{\textcolor{red}{\overrightarrow{\Pi}(.|X)}\sim \widetilde{Z}}{\underset{X\sim \overleftarrow{\Pi}(.|\widetilde{Z})}\leftrightarrows}& \Phi
\\[1cm]
\Psi&
 \overset{\widetilde{X}\sim\overleftarrow{\Pi}(.|Z)}{\underset{\overrightarrow{\Pi}'(.|\widetilde{X})\sim Z}\leftrightarrows} &
Q
\end{array}
\]
\end{subfigure}
\unskip\ \vrule\ 
\begin{subfigure}[b]{0.45\textwidth}
\[
\begin{array}{ccc}
x-obs. && z-lat. \\
P &\overset{\overrightarrow{\Pi}(.|X)\sim \widetilde{Z}}{\underset{X\sim \overleftarrow{\Pi}'(.|\widetilde{Z})}\leftrightarrows}& \Phi
\\[1cm]
\Psi&
 \overset{\widetilde{X}\sim\textcolor{red}{\overleftarrow{\Pi}(.|Z)}}{\underset{\overrightarrow{\Pi}(.|\widetilde{X})\sim Z}\leftrightarrows} &
Q
\end{array}
\]
\end{subfigure}

\caption{TMC approach (left) compared to DIF (right) - parameterized transports are indicated in red}
\end{figure}
\\
To summarize, the TMC approach considers a forward transport $\overrightarrow{\Pi}$ between $P$ and $\Phi$, then computes its backward transport $\overleftarrow{\Pi}$ which is finally applied to $Q$ in order to obtain the model $\Psi$. DIF and TMC are in fact defined in reverse order compared to one another since in the DIF approach, we consider a backward transport $\overleftarrow{\Pi}$ between $Q$ and $\Psi$, and we consequently deduce the forward transportation $\overrightarrow{\Pi}$ which can then be applied to $P$ in order to obtain $\Phi$. As a consequence, though the expressions for $\psi(x), \phi(z), \overleftarrow{\pi}(x|z)$ and $\overrightarrow{\pi}(z|x)$ look similar and are obtained via similar computations, in the TMC approach we set $v_k$ and compute $w_k$ while in the DIF approach we set $w_k$ and compute $v_k$.

In section \ref{DIFVI} we will discuss the pros and cons of each approach under the scope of the VI problem, 
but at this point 
let us already notice that TMC cannot be applied to a density estimation setting. 
Indeed, 
with this choice of parameterization,
$w_k$ given in \eqref{w_k_TMC} is computed from $v_k$ but also depends on the density $p$ which is not available in the density estimation setting. 
On the other hand,
in the DIF approach, functions $w_k$ are parameterized directly and do not depend on pdf $p$. Consequently, we can compute directly $w_k$ and $\psi(x)$, which enables sampling and density evaluation in both settings of VI and VDE. Finally, we can already argue in favor of DIF since it provides with a more versatile tool for tackling both variational problems.

\subsection{DIF for VI}\label{DIFVI}

So far, we have presented the general principles of DIF and explained that it consists in a natural extension of NF. In particular, even though the transformation is now stochastic, DIF defines a model pdf $\psi$ which remains computable, and also provides the ability to go back and forth between observed and latent spaces. 

In this section we discuss the use of a DIF for tackling a VI problem and we therefore consider the setting described in \ref{VI}. As we have mentionned in the previous section \ref{TMC_approach}, 
DIF and TMC are closely related. However, TMC considers a particular optimization objective function. In section \ref{Computational-aspects} and \ref{Variational-aspects} we further compare TMC to DIF in order to discuss the relevance of this optimization objective, 
and finally in section \ref{RBVI} we argue in favor of a more motivated optimization objective.

\subsubsection{Computational aspects}
\label{Computational-aspects}

With the same notations introduced before, TMC builds a model $\Psi$ for $P$ by solving the optimization problem:
\begin{equation}\label{maxi_tmc}
    \max_{\theta \in \Theta} \sum_{\stackrel{i=1}{z_i\sim Q}}^{M} \log\paren{\phi(z_i)},
\end{equation}
which corresponds to minimizing
an MC approximation of $D_{\mathrm{KL}}\paren{Q||\Phi}$. 

However, the optimization problem \eqref{maxi_tmc}
seems strange at first sight, 
because the standard approach for VI would indeed prescribe 
minimizing a discrepancy between $\Psi$ and $P$ (see \eqref{DKL_VI}).
But minimizing an MC approximation of $D_{\mathrm{KL}}\paren{\Psi||P}$ would yield the optimization objective: 
\begin{equation}
    \label{tmc-optim-normal}
    \min_{\theta \in \Theta} \sum_{\stackrel{i=1}{x_i\sim \Psi}}^{M} \log\paren{\frac{\psi(x_i)}{p(x_i)}}.
\end{equation}
Since the samples $x_i \sim \Psi$ depend on model parameters $\theta$, 
we should apply a reparameterization trick, 
that is,
write $x_i$ as $x_i = T^{-1}(\epsilon_i; \theta)$ 
in which $\epsilon_i \sim \epsilon$ and 
rv $\epsilon$ does not depend on $\theta$.
In the case of NF (see section \ref{appli1}),
the deterministic mapping automatically induced a differentiable reparameterization trick $x_i = T^{-1}(z_i)$. 
Here by contrast, 
since sampling from $\Psi$ 
involves sampling from an auxiliary categorical latent variable, 
finding an invertible change of variable
which is differentiable wrt $\theta$ is likely not to be possible. Therefore, the minimization 
problem \eqref{tmc-optim-normal} cannot be conducted via GD.

By contrast, 
the objective function in \eqref{maxi_tmc} 
is easy to maximize: 
sampling from $Q$ is straightforward by design, 
and this objective is differentiable with respect to $\theta$ and can therefore be maximized via GA.
This computational argument argues in favor of the optimization objective in the TMC approach, 
but on the other hand, one can wonder whether 
minimizing a discrepancy in the latent space 
induces similar counterparts in the observed space,
see section \ref{Variational-aspects} below.
\subsubsection{Variational aspects}
\label{Variational-aspects}

Unfortunately, by contrast with NF, 
the two equalities between $D_{\mathrm{KL}}$ \eqref{DKLEqualityReverseObserved} and \eqref{DKLEqualityForwardObserved}
no longer hold when we work with a DIF or a TMC model. Therefore it is not so obvious that minimizing a $D_\mathrm{KL}$ in the latent space, that is between $Q$ and $\Phi$, produces a good approximation of $P$ with $\Psi$. Nonetheless, we can justify to some extent the use of this optimization objective in TMC. Indeed, we notice that the forward $D_\mathrm{KL}$ in the latent space is an upper bound of the reverse $D_\mathrm{KL}$ in the observed space:
\begin{align*}
    D_\mathrm{KL}\paren{Q||\Phi}  &
    = D_\mathrm{KL}\paren{\Psi||P} + \E_{X\sim\Psi}\bracket{D_\mathrm{KL}\paren{\overleftarrow{\Pi}(.|X)||\overleftarrow{\Pi}'(.|X)}} \geq D_\mathrm{KL}\paren{\Psi||P}.
\end{align*}
Note moreover that we have similarly $D_\mathrm{KL}\paren{\Phi||Q} \geq D_\mathrm{KL}\paren{P||\Psi}$. Therefore, the TMC approach minimizes (an MC approximation of) an upper bound of the usual optimization objective \eqref{DKL_VI} defined in the VI setting. 

Since $D_\mathrm{KL}$ are positive, it follows that if a $D_{\mathrm{KL}}$ between $\Phi$ and $Q$ (forward or reverse) reaches zero via optimization, both forward and reverse $D_{\mathrm{KL}}$ between $\Psi$ and $P$ reach zero. 
However, 
forcing a $D_{\mathrm{KL}}$ in the latent space to zero
means that the prior pdf $q$ belongs to $\set{\phi|\theta \in \Theta}$,
or equivalently that there exists  
$\overleftarrow{\Pi}$ and $q$ such that 
$\mathcal{F}\paren{q;\overleftarrow{\Pi}} = p$
(see \eqref{fonctionnelle-DIF}),
which is unlikely to be the case for arbitrary distributions $P$.
Moreover, 
standard optimization techniques such as GD only guarantee convergence to a local extremum of the objective function. So in practice we have to deal with positive $D_{\mathrm{KL}}$ in the latent space as we may only reach a local minimum of a positive function.
There is furthermore no evidence that a local minimum of $D_\mathrm{KL}$ in the latent space is also a local minimum of $D_{KL}$ in the observed space. Finally we cannot conclude with certainty 
that a decent approximation of $Q$ with $\Phi$ (in the $D_{\mathrm{KL}}$ sense)
produces a good model $\Psi$ and we would preferably want to obtain a minimum of $D_\mathrm{KL}$ in the observed space. \newline

In the case of the DIF, since the model is defined the other way round as compared to TMC,
the majorization obtained for TMC becomes a minorization: 
\begin{align*}
    D_\mathrm{KL}\paren{\Psi||P}  &
    = D_\mathrm{KL}\paren{Q||\Phi} + \E_{Z\sim Q}\bracket{D_\mathrm{KL}\paren{\overrightarrow{\Pi}(.|Z)||\overrightarrow{\Pi}'(.|Z)}} \geq D_\mathrm{KL}\paren{Q||\Phi}. 
\end{align*}
Therefore, in the case of DIF, minimizing a latent $D_\mathrm{KL}$ would only minimize a lower bound
of the observed $D_{KL}$ 
which we should minimize in the VI setting. 
This consideration would argue against DIF
if we were not able to minimize directly the $D_{KL}$ in the observed space. 
Fortunately, 
as we now see,
it is indeed possible 
in both the DIF and TMC cases
to minimize directly an MC approximation of the $D_{KL}(\Psi||P)$. 

\subsubsection{Rao-Blackwellizing the $D_{\mathrm{KL}}$ estimate}\label{RBVI}
From sections 
\ref{Computational-aspects} and
\ref{Variational-aspects},
we see that it would be desirable to minimize a discrepancy measure in the observed space. Could we write an estimate of $D_{\mathrm{KL}}(\Psi||P)$ for which we are able to compute the gradients wrt model parameters $\theta$ ?
Such a case would be ideal, since we could perform GD while ensuring that the model converges toward a local minimum of the discrepancy measure in the observed space.

It happens that the following estimator : 
\begin{equation}\label{RaoBlackwellDKL}
    D_{\mathrm{KL}}\paren{\Psi||P} \approx \frac{1}{M}\sum_{\stackrel{i=1}{z_i\sim Q}}^{M}\sum_{k=1}^K w_k(z_i)\log\paren{\frac{\psi\paren{T_k^{-1}(z_i)}}{p\paren{T_k^{-1}(z_i)}}}
\end{equation}
is one possible solution
since,
by contrast with \eqref{tmc-optim-normal}, this $D_{\mathrm{KL}}$ estimate is differentiable with respect to model parameters.
As we now see, 
this estimate indeed comes as the result 
of a Rao-Blackwellization (RB) procedure \citep{RaoBlackwell} \citep{gelfand-smith} \citep{RB_MCMC}.
Let $J$ be the rv
$$
J =\log\paren{\frac{\psi\paren{\widetilde{X}}}{p\paren{\widetilde{X}}}}
 \stackrel{\eqref{DIF-chgt-de-variable}}{=}
 \log\paren{\frac{\psi\paren{T_U^{-1}(Z)}}{p\paren{T_U^{-1}(Z)}}} ,
$$
where
$Z\sim Q$
and 
$U \sim \textrm{Categorical}\paren{w_1(Z),..., w_K(Z)}$.
It is clear that 
$$
D_{\mathrm{KL}}(\Psi||P) =\E\paren{J} ,
$$
where the expectation is taken wrt the joint distribution of $\paren{Z,U}$. 
On the one hand, 
sampling $\set{(z_i, u_i)}_{i=1,...,M}$
from this joint distribution
yields the crude MC estimate in \eqref{tmc-optim-normal}.  On the other hand, RB is based on the observation that $\E\paren{J} = \E\paren{\E\paren{J|Z}}$ \citep{RaoBlackwellTheorem}.
Since we can compute the inner expectation: 
\begin{align*}
    \E\paren{J|Z} = \sum_{k=1}^K w_k(Z)\log\paren{\frac{\psi\paren{T_k^{-1}(Z)}}{p\paren{T_k^{-1}(Z)}}},  
\end{align*}
only the outer one calls for an MC approximation,
so we only need to sample $z_i \sim Q$.
This leads to the estimator 
$D_{\mathrm{KL}}(\Psi||P) \approx \frac{1}{M} \sum_{i=1}^M \E\paren{J|Z=z_i}$,
which is nothing but \eqref{RaoBlackwellDKL}. 

The interest of using this RB estimate is twofold.
First, 
as is well known
$\Var\paren{\E\paren{J|Z}}  = \Var\paren{J} - \mathbbm{E}\paren{\Var\paren{J|Z}}$,
so \eqref{RaoBlackwellDKL}
has lower variance than the estimator in \eqref{tmc-optim-normal}. 
Next (and more importantly in the context of this paper),
we are no longer reliant on a reparameterization of the Categorical rv $U$, 
so the estimate is now differentiable. 
Indeed, before resorting to an MC approximation, we have computed whatever could be computed, namely $\E\paren{J|Z}$ (where the expectation is taken with respect to $U$). Therefore the estimate does not involve sampling from $U$ since this rv has been explicitly marginalized out.

\subsection{DIF for VDE}\label{DIFDE}

As we have seen already,
DIF are designed such that they can also be used for VDE, 
since the backward transport and density $\psi$ do not depend on $p$ (remember that $p$ is not available in a density estimation setting, see section \ref{DE}).
In this section we now explain precisely
how DIF can be used for the VDE problem.

\subsubsection{The MLE approach}

As we now see, 
the issues that were identified when using DIF for VI no longer occur when tackling the problem of VDE with a DIF model. 
To see this, suppose that we dispose of samples $x_i \sim P$. 
Since $\psi$ can be written in closed form,
the MLE problem reads
\begin{equation}\label{DIFObjective}
    \max_{\theta \in \theta} \sum_{\stackrel{i=1}{x_i\sim P}}^{M} \log\paren{\psi(x_i)}
    \stackrel{\eqref{DIFDensity}}{=}
    \max_{\theta \in \theta} \sum_{\stackrel{i=1}{x_i\sim P}}^{M} \log\paren{\sum_{k=1}^K w_k\paren{T_k(x_i)}q\paren{T_k(x_i)}\abs{\det \mathrm{J}_{T_k}(x_i)}} .
\end{equation}
By contrast with
\eqref{tmc-optim-normal}, $x_i$ are sampled from $P$, and not from $\Psi$,
so do not depend on $\theta$.
As a result,
we see from the rhs of \eqref{DIFObjective} that  this objective is
differentiable wrt $\theta$.

\subsubsection{A Generalized Expectation-Maximization (GEM) procedure}\label{GeneralizedEMApproach}

Let us turn to computational optimization aspects.
Of course,
the objective function in \eqref{DIFObjective} 
can be directly optimized with GD in an Automatic Differentiation framework like Pytorch or Tensorflow. However, let us observe that this function involves the logarithm of a sum, which leads to entangled gradients.
As a consequence, gradients computation could be slowed down.

In this section we propose an alternative optimization procedure, 
based on a Minorize-Maximization \citep{MajorationMinoration} approach,
the principle of which is as follows. Instead of optimizing directly a function 
$f(\theta)$,
we sequentially build a series of surrogate functions 
$\set{g_{ \theta_1}(\theta), g_{ \theta_2}(\theta), \cdots}$
which locally minorate $f$ and for which $g_{ \theta_t}(\theta_t) = f(\theta_t)$.
From this series of functions, one can sequentially deduce a series of parameters $\set{\theta_1,\theta_2,...}$ such that $g_{ \theta_t}(\theta_{t+1}) \geq g_{ \theta_t}(\theta_t)$.
So
\begin{equation*}
    f(\theta_{t+1}) \geq g_{\theta_t}(\theta_{t+1})\geq g_{\theta_t}(\theta_t) = f(\theta_t),
\end{equation*}
which finally ensures that $f(\theta_t)$ converges to a local maximum of $f(\theta)$.
Moreover, if both $f(\theta)$ are $g_{\theta_t}(\theta)$ are differentiable functions, then the gradients evaluated at $\theta = \theta_t$ must be equal. To see this, let us consider the function $\theta \mapsto f(\theta) - g_{\theta_t}(\theta)$; in the vicinity of $\theta_t$, this function is non-negative, differentiable and is zero for $\theta = \theta_t$. Hence $\theta_t$ is a local minimum and its gradient is zero: 
\begin{equation}\label{egalité_gradients}
    \nabla_{\theta}f\paren{\theta}\rvert_{\theta = \theta_t} = \nabla_{\theta}g_{\theta_t}\paren{\theta}\rvert_{\theta = \theta_t}.
\end{equation}

We now apply this technique to the above optimization problem.
We obtain the surrogate function
$g_{\theta_t}(\theta)$ for the likelihood function in
\eqref{DIFObjective}
(we omit variables $x_i$ in the lhs of \eqref{GEM_surrogate} since the samples are fixed);
details are given in Appendix \ref{GradientEM}:
\begin{eqnarray}
    \label{GEM_surrogate}
    g_{\theta_t}(\theta)
    &=&
    \sum_{i=1}^M\sum_{k=1}^K v_k^{(\theta_t)}(x_i)\log\paren{\frac{h_k^{(\theta)}(x_i)}{v_k^{(\theta_t)}(x_i)}} , \\
    \nonumber
    h_k^{(\theta)}(x_i) 
    &=&
    w_k^{(\theta)}\paren{T_k^{(\theta)}(x_i)}q\paren{T_k^{(\theta)}(x_i)}\abs{\det \mathrm{J}_{T_k^{(\theta)}}(x_i)} .    
\end{eqnarray}
If we perform GA with respect to $\theta$
(see Algorithm \ref{gradientEMalgo} below), we obtain a GEM scheme \citep{ConvergenceEM}, which ensures that the model converges toward a local maximum in \eqref{DIFObjective}. 
Finally observe that our surrogate function no longer involves a log-sum but rather a sum-log, 
which detangles the computation of its gradients.

\RestyleAlgo{ruled}
\begin{algorithm}[h!]
\caption{maximization of log-likelihood via GEM}\label{gradientEMalgo}
\KwData {data $\set{x_1,...,x_M}$, $\theta_0$ initial parameters, $T$ No. steps, $\eta$ learning rate}
\KwResult {$\theta_T$}
    \For{t=0 to T-1}{    
        $\theta_{t+1} = \theta_t + \eta\paren{\nabla_{\theta}g_{\theta_t}\paren{\theta}\rvert_{\theta = \theta_t}}$
    }
\end{algorithm}

\section{DIF in practice}\label{dif_in_practice}

In this section we first explain how one can parameterize the functions $T_k$ and $w_k$ to produce an efficient DIF model. 
We then propose an overall view of the DIF mechanism, as a stochastic transport transforming $Q$ into the resulting distribution $\Psi$.
We then revisit DIF as an extension of mixture density models but where the constant weights are replaced by an arbitrary function of $z$; we illustrate this effect on complex distributions,
and see that DIF enable to capture sharp edges and finer details as compared to standard Gaussian Mixture Models.
We also see that complex DIF can be constructed as the succession of simpler building blocs: in the same spirit as NF,
we propose an approach for cascading DIF layers. 
Finally we discuss the use of DIF in the specific setting of conditional density estimation,
and see that one could easily turn a DIF into a conditional density model.

\subsection{Example of DIF parameterization}\label{Parameterization}

As we have explained before, 
using DIF requires solving an optimization problem 
(be it for the VI or VDE problems), 
in which the multidimensional parameter $\theta$ 
gathers those of
the probability functions $w_k$, 
as well as of the invertible mappings $T_k$.
Even though this  optimization is performed wrt the parameters altogether,
$w_k$ and $T_k$ still play a different role and must be specified accordingly. 

\subsubsection{Probability functions $w_k$}\label{probabilitypartitioning}
The functions $w_k(.)$ are straightforward to parameterize, since the only constraints are that these functions are differentiable, non negative, and sum to 1 for any given input vector $z$. 
This can be achieved by defining 
$w_1(z),...,w_K(z)$ as the output of a $K$-label classifier architecture with input $z$ by computing the unormalized weights $\widetilde{w_1}(z),...,\widetilde{w_K}(z)$ and applying a softmax normalization. This ensures that the weights sum to one and form a valid vector of categorical probabilities.
More precisely, we can for instance consider an NN function with $L$ hidden layers, each layer $l = 1,...,L$ having $n_l$ hidden units: 
\begin{align*}
    & h_1 = \sigma(W_0 z + b_0); \\
    & h_{l+1} = \sigma(W_l h_l + b_l) \text{ for } l = 1,...,L-1 ;\\
    & \bracket{\widetilde{w_1}(z),...,\widetilde{w_K}(z)}^{\mathsmaller T} = W_L h_{L} + b_L; \\
    & \bracket{w_1(z),...,w_K(z)}^{\mathsmaller T}= \mathrm{Softmax}\paren{\bracket{\widetilde{w_1}(z),...,\widetilde{w_K}(z)}^{\mathsmaller T}},
\end{align*}
where $W_l \in \mathbbm{R}^{n_{l+1}\times n_{l}}$ and $b_l \in \mathbbm{R}^{n_{l+1}}$ for $l=0,...,L$ (with $n_0 = d$ and $n_{L+1} = K$) are the weights and biases parameters, and where $\sigma(.)$ is some chosen element-wise activation function (for example the sigmoid function). It turns out that

\subsubsection{Invertible maps - $T_k$}
\label{invertible-map}
When selecting the parametric functions $T_k$, we actually dispose of a wide range of possibilities. 
Depending on the problem,
the only constraint is that the functions must be changes of variables, 
and that $T_k^{-1}$ (for VI)
or 
the Jacobian Determinant 
(for VDE)
can be computed easily.
We may consider simple location-scale mappings like in Gaussian Mixture Models (GMM), or we may borrow from the NF literature such as in \citep{NICE} \citep{RealNVP} \citep{MAF} \citep{IAF} \citep{Glow} \citep{NeuralSplineFlows}. In section \ref{mixed_models} we propose a construction which reduces the burden of parameterizing the mappings $T_k$. Moreover, if we consider weights $w_k$ defined via a flexible parametric function (as in section \ref{probabilitypartitioning}), we do not require flexible invertible mappings $T_k$ to produce a flexible DIF. 
Therefore we only consider here a simple location-scale
\begin{equation*}
    T_k^{-1}(z) = \mu_k + s_k\odot z,
\end{equation*}
where $\mu_k \in \mathbbm{R}^d$ 
(a translation vector, whence the term \emph{location}) 
and $s_k\in {\mathbbm{R}_+^*}^d$ 
(a \emph{scale} vector)
are the parameters be optimized,
and $\odot$ is the element wise vector product. Since $s_k$ is strictly positive, $T_k$ is invertible and we can easily obtain the inverse mapping as well as the Jacobian determinant with: 
\begin{equation*}
    T_k(x) = s_k^{-1}\odot(x- \mu_k) \text{ and }  \abs{\det \mathrm{J}_{T_k}(x)} = \prod_{j = 1}^d\bracket{s_k}_{j}^{-1},
\end{equation*}
where $s_k^{-1}$ is the vector of element-wise inverses of $s_k$. 

\subsection{An overall view: the DIF de- and re-constructs $Q$ into $\Psi$}
\label{decortiquer}

\begin{figure}
\centering
\begin{subfigure}[b]{.8\linewidth}
\includegraphics[width=\linewidth]{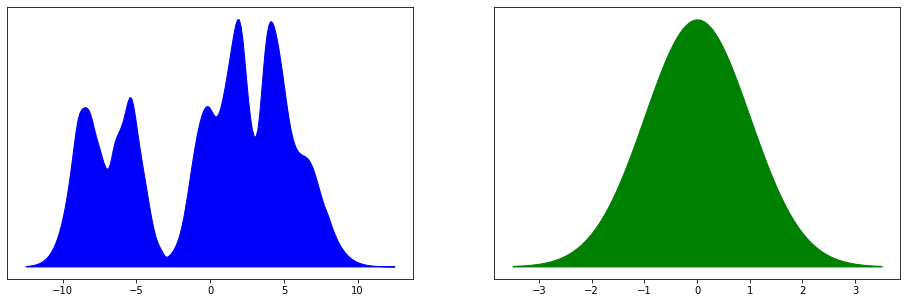}
\setcounter{subfigure}{0}%
\caption{A DIF transforms the prior pdf $Q$ (green) into the final pdf $\Psi$ (blue)}\label{fig:total}
\end{subfigure}
\begin{subfigure}[b]{.8\linewidth}
\includegraphics[width=\linewidth]{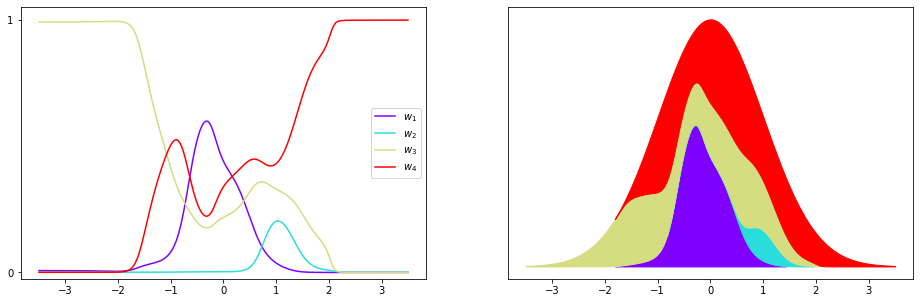}
\setcounter{subfigure}{1}%
\caption{Partitioning of the latent space with functions $w_k(z)$}\label{fig:partitioning}
\end{subfigure}
\begin{subfigure}[b]{.8\linewidth}
\includegraphics[width=\linewidth]{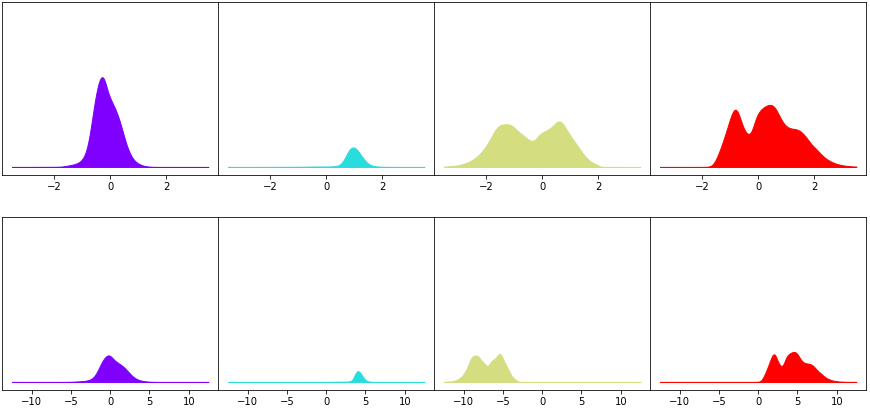}
\setcounter{subfigure}{2}%
\caption{Elements of prior mass transported with mappings $T_k^{-1}$}\label{fig:transport}
\end{subfigure}
\begin{subfigure}[b]{.45\linewidth}
\includegraphics[width=\linewidth]{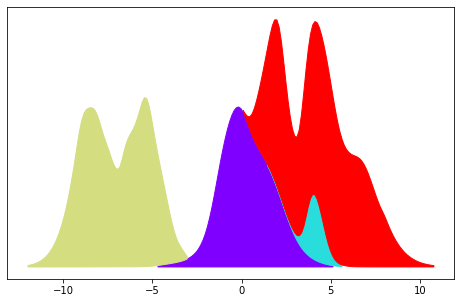}
\setcounter{subfigure}{3}%
\caption{Recombination}\label{fig:recombination}
\end{subfigure}
\caption{DIF mechanism for decomposing / recomposing $Q$ into $\Psi$}\label{fig:dif_example}
\end{figure}
We finally illustrate via a one-dimensional example how a DIF transforms 
the prior distribution $Q$
into a
complex probability distribution $\Psi$ with density given by \eqref{DIFDensity}
(see figure \ref{fig:total}), 
and in particular explain the roles of weights $\set{w_k}_{k=1,...,K}$ and of mappings 
$\set{T_k}_{k=1,..., K}$. 
The discussion in this section is of course independent of the problem tackled (VI or VDE) and of the associated optimization. 

First, lhs of figure \ref{fig:partitioning} displays 
the weight functions $w_k(z)$.
Since they are positive and sum to $1$ (for any $z$),
we have 
$q(z) = \sum_{k=1}^K w_k(z) q(z)$;
so functions $w_k(z)$ 
induce a soft partitioning of the latent space, and indeed split the prior mass into several parts, 
see rhs of figure \ref{fig:partitioning} 
(or equivalently the first row of figure \ref{fig:transport}).
These figures indeed provide
a way of visualizing the joint distribution $(Z, U)$: 
the values of $z$ can be read on the $x$-axis, 
and the values of $U=1,...,K$ 
are the different colors in the rhs of figure \ref{fig:partitioning}
(or the different sub-figures in figure \ref{fig:transport}).

Next, given $U=k$, a prior sample $z$ is transported via mapping $T_k$ with $x = T_k^{-1}(z)$. 
So on the whole,
the C1-diffeomorphisms $T_k^{-1}$
send the elements of mass in possibly different regions of the observed space,
and continuously reshape them
(see second row of figure \ref{fig:transport}).

Finally all  
these parts are recombined into the final probability distribution $\Psi$, see figure \ref{fig:recombination}.

\subsection{DIF vs Mixture Models}\label{DIFvsMixture}
So far we have presented DIF as a non-deterministic extension of deterministic NF. However, as already mentioned in section \ref{DIF_as_LVM}, due to their discrete latent structure, DIF can be connected to mixture models as well. 
Indeed we retrieve a mixture model when considering a DIF with constant $w_k$ functions.
In this section we discuss the pros and cons of DIF (that is, weights depend on the latent value $z$) as opposed to mixtures models
(that is, with constant weights).

First, as is displayed in section \ref{decortiquer}, 
a DIF is particularly well suited for capturing multimodality, 
because two phenomenons add up: 
just like in mixture models, the elements of mass are dispatched into several regions of space; but these elements themselves can be turned multimodal,
since the prior $q$ is reshaped by a function $w_k(z)$. 
For instance in figure \ref{fig:dif_example}, 
a distribution with 5 modes
was captured with only $K=4$ components.

\begin{figure}[!h]
    \centering
    \begin{subfigure}[b]{0.24\textwidth}
        \includegraphics[width=\textwidth]{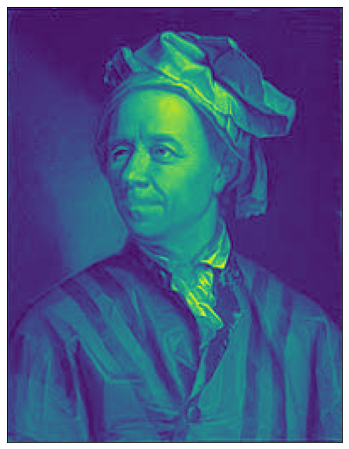}
    \end{subfigure}
    \begin{subfigure}[b]{0.24\textwidth}
        \includegraphics[width=\textwidth]{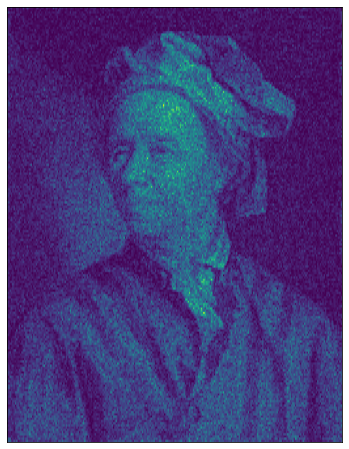}
    \end{subfigure}
    \begin{subfigure}[b]{0.24\textwidth}
        \includegraphics[width=\textwidth]{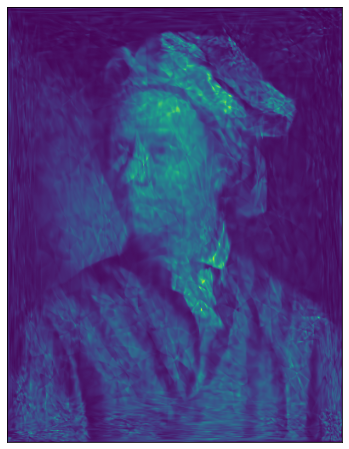}
    \end{subfigure}
    \begin{subfigure}[b]{0.24\textwidth}
        \includegraphics[width=\textwidth]{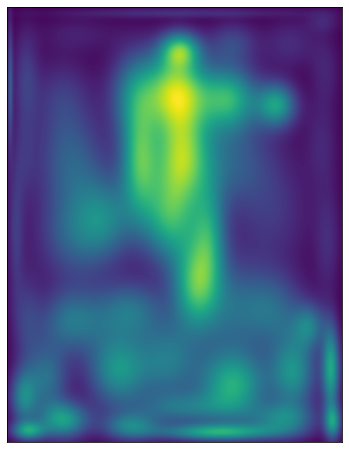}
    \end{subfigure}
    \centering
    \begin{subfigure}[b]{0.24\textwidth}
        \includegraphics[width=\textwidth]{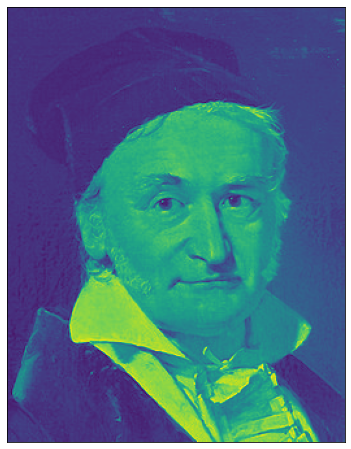}
    \end{subfigure}
    \begin{subfigure}[b]{0.24\textwidth}
        \includegraphics[width=\textwidth]{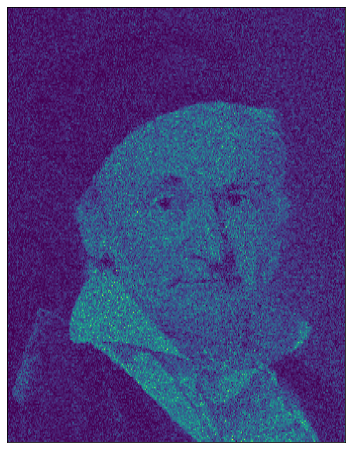}
    \end{subfigure}
    \begin{subfigure}[b]{0.24\textwidth}
        \includegraphics[width=\textwidth]{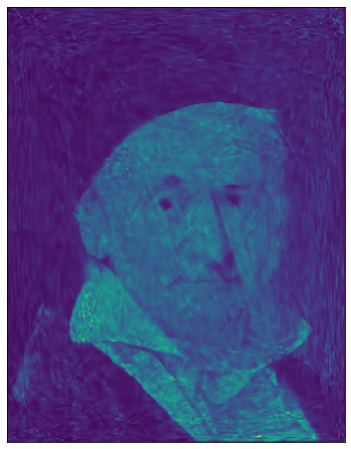}
    \end{subfigure}
    \begin{subfigure}[b]{0.24\textwidth}
        \includegraphics[width=\textwidth]{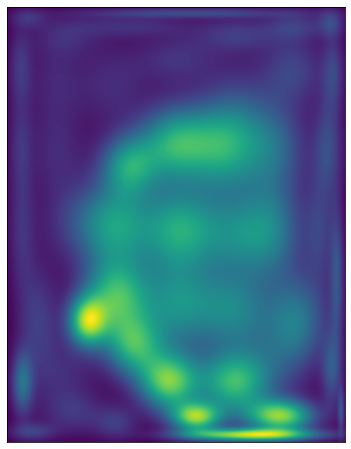}
    \end{subfigure}
    
    \centering
    \begin{subfigure}[b]{0.24\textwidth}
        \includegraphics[width=\textwidth]{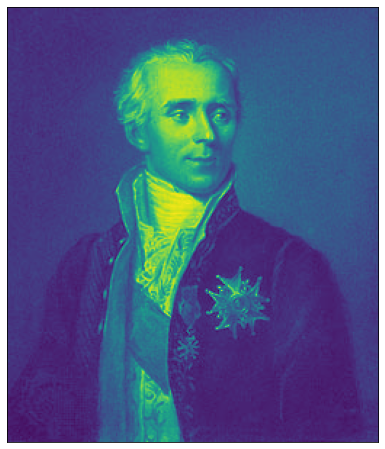}
    \end{subfigure}
    \begin{subfigure}[b]{0.24\textwidth}
        \includegraphics[width=\textwidth]{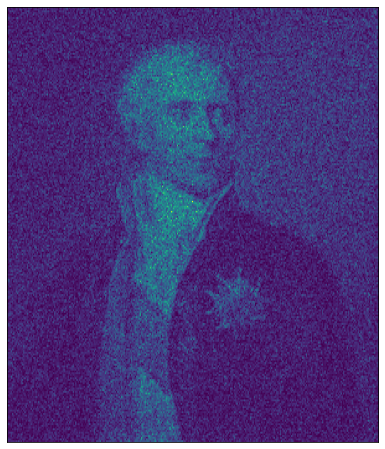}
    \end{subfigure}
    \begin{subfigure}[b]{0.24\textwidth}
        \includegraphics[width=\textwidth]{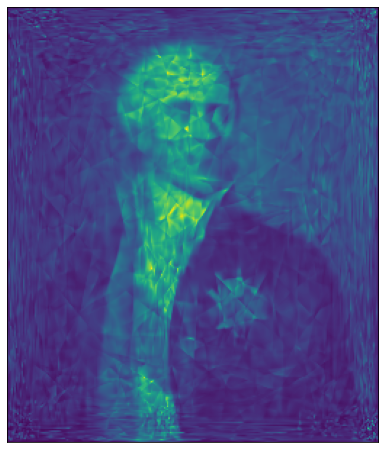}
    \end{subfigure}
    \begin{subfigure}[b]{0.24\textwidth}
        \includegraphics[width=\textwidth]{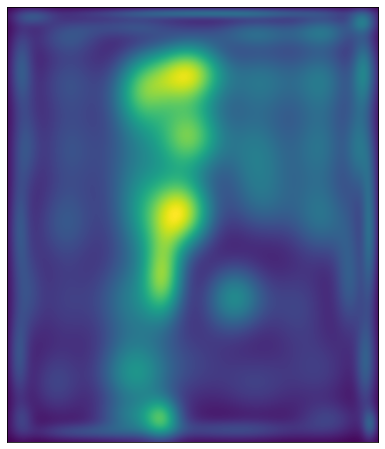}
    \end{subfigure}
    \caption{A DIF (middle-right) can approach the distribution associated to an image (left) from samples (middle-left) - compared to a GMM obtained via EM (right).}\label{fig:image_dif}
\end{figure}

Second, flexible probability functions $w_k(z)$ as those proposed in section \ref{probabilitypartitioning}
enable to reach distributions with sharp edges, 
and even close to discontinuous pdf.
This expressivity is illustrated in figure \ref{fig:image_dif} where we consider the VDE setting 
(similar conclusions would apply in the VI setting). 
In this example, we treat the Euler, Gauss and Laplace grey-scale images (first column) 
as 2-dimensional simple functions, 
and thus as 2-dimensional pdf with values proportional to the intensity of a pixel. 
We next sample from these distributions (second column).
We finally proceed to VDE,
either by using a DIF with $K=40$ components 
and with weights $w_k(z)$ defined via an NN function with 3 layers,
each with 128 hidden units (third column);
or a GMM with the same number $K=40$ of components
(fourth column).
We see from these qualitative examples that, as expected, the GMM is not precise enough to capture all the details and variations in the images,
while by contrast DIF can efficiently represent the distributions with their fine details, sharp edges and discontinuities. 

On the other hand, 
in the specific VDE setting,
a drawback of DIF as compared to GMM
is that we no longer dispose of 
an efficient optimization procedure. 
More precisely, 
when the weights are constant, 
the maximum of \eqref{GEM_surrogate}
can be computed in closed-form, 
which yields the well-known Expectation-Maximisation (EM) procedure for GMM \citep{EMAlgorithm}.
DIF no longer benefit from the same advantage, 
since neither the log-likelihood function \eqref{DIF_likelihood} nor the surrogate function \eqref{GEM_surrogate} admit closed form maxima,
and we therefore can only resort to gradient-based optimisation procedures. 
However, as a rescue we can at least
improve the training of a DIF model by  starting from an initial state obtained by using an EM procedure in a GMM model.
Of course, this applies only with location scale parameterization $T_k^{-1}(z) = \mu_k + {\Sigma_k}^{\frac{1}{2}}z$ where $\Sigma_k$ is a covariance matrix (possibly diagonal, as in section 
\ref{invertible-map}). This can be achieved effortlessly by simply initializing to zero $W_L$ (the weights of the last hidden layer which computes $w_k$), 
and setting $b_L$, $\mu_k$, $\Sigma_k$ respectively to the mixture weights, means and covariances retrieved by the EM procedure.

\subsection{Cascading DIF}\label{mixed_models}

We now see that, in same spirit as NF, we can cascade simple DIF together in order to produce expressive models.  

\subsubsection{Methodology}\label{methodology_cascade}

Remember from section \ref{whyNF} that
NF can be constructed as a composition of successive transforms.
As we now see, it is also possible to cascade DIF themselves:
stacking two (or more) DIF produces a DIF,
so DIF can be used as elementary building blocks
for defining elaborate models and transforms.
To see this, 
consider the following cascade of two DIF $\overleftarrow{\Pi}^{[0]}$ and $\overleftarrow{\Pi}^{[1]}$: 
 \[
\begin{array}{cccccccccc}
x && z_1 && z \\
P & 
\xrightarrow{\overrightarrow{\Pi}^{[0]}(x)}&
\Phi_1 &\xrightarrow{\overrightarrow{\Pi}^{[1]}(z_1)}& \Phi
\\[0.2cm]
\Psi & 
\xleftarrow{\overleftarrow{\Pi}^{[0]}(z_1)}&
\Psi_1 &\xleftarrow{\overleftarrow{\Pi}^{[1]}(z)}& Q
\end{array}
\]
It is easy to see that the equivalent backward transportation is given by: 
\begin{eqnarray}
\label{compose-DIF1}
    \overleftarrow{\pi}^{[0,1]}(x|z) & = &
    \sum_{k_1=1}^{K_1} \sum_{k_0=1}^{K_0} w_{k_0,k_1}(z) \delta_{T_{k_0,k_1}^{-1}(z)}(x), \\
    \label{compose-DIF2}
    T_{k_0,k_1}(x) & = & {T_{k_1}^{[1]}}\paren{{T_{k_0}^{[0]}}(x)}, \\
    \label{compose-DIF3}
    w_{k_0,k_1}(z) & = & w_{k_0}^{[0]}\paren{{T_{k_1}^{[1]}}^{-1}(z)}w_{k_1}^{[1]}(z).
\end{eqnarray}
Comparing \eqref{compose-DIF1} with \eqref{DIF_likelihood},
we see that $\overleftarrow{\Pi}^{[0,1]}$ is indeed a DIF 
with $K_1 \times K_0$ components; the probability mass is split into $K_1 \times K_0$ components, 
but by using only the equivalent of $K_1 + K_0$ parameters.

This construction generalizes the discussion in section \ref{whyNF}
(which corresponds to the case $K_0 = K_1 = 1$),
and includes as particular cases the cascading of DIF with NF
($K_0 = 1$ or $K_1 = 1$).
Moreover,
we see from \eqref{compose-DIF2} and \eqref{compose-DIF3} that apart from an increased number of components
(which are correlated since they share a smaller set of parameters),
cascading DIF potentially enables to create elaborate mappings
from simple ones.

In particular, cascading a DIF ($K_1 > 1$) with simple mappings ${T_{k_1}^{[1]}}$ (such as location-scale) with a NF ($K_0 = 1$) in which mapping $T$ is a flexible change of variables (such as \citep{NICE} \citep{RealNVP} \citep{MAF} \citep{IAF} \citep{Glow} \citep{NeuralSplineFlows}) produces a new DIF with $K_1$ components,
but with more flexible mappings than those used in the initial DIF. 
Of course, the discussion in this section can be extended to more than two DIF as the principles apply recursively. 
Finally, cascading DIF induces a particular expression of the $D_{\mathrm{KL}}$ to be minimized (both for the VI and the VDE problems), see Appendix \ref{cascade-practice} for details.

\subsubsection{Breaking topological limitation of NF}
\label{breakbarrier}

Remember from section
\ref{decortiquer} that the prior probability mass is split into several components,
enabling DIF to express multimodal pdf and/or pdf with disjoint support.
So DIF can be used for breaking the topological limitation evoked in section \ref{topological}.

\begin{figure}
\centering
\includegraphics[width=\linewidth]{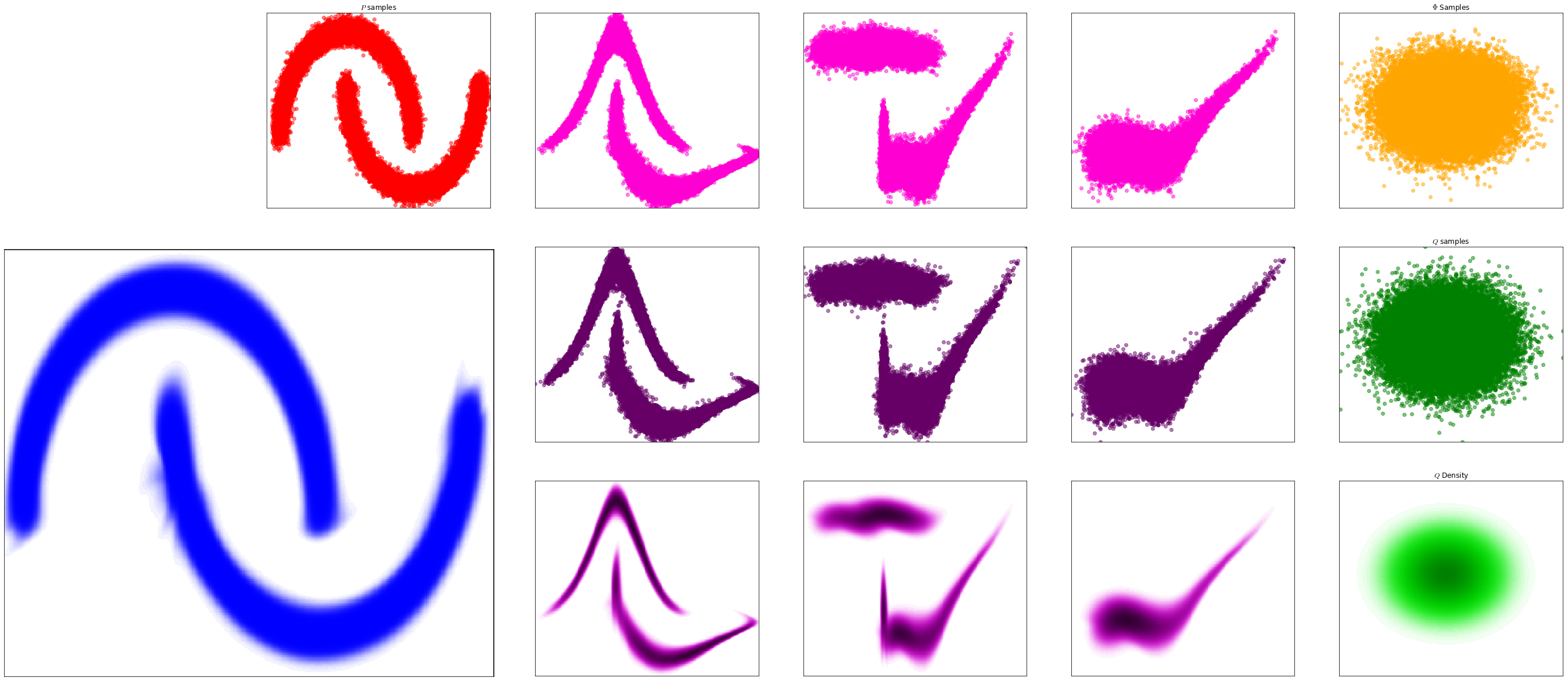}
\caption{Breaking the topological barrier using a DIF within a NF}\label{fig:BreakingTopo}
\end{figure}

Let us illustrate this via the following example. 
In figure \ref{fig:BreakingTopo} we display how a Gaussian prior can be transformed simply into a pdf with disjoint support.
This is achieved by a DIF which was built as a cascade, as explained in section \ref{methodology_cascade}.
More precisely,
the only difference with figure \ref{fig:constraint} is that a DIF was included in the flow steps (in between the 2nd and the 3rd).
This DIF element was purposely chosen to be simple with $K=2$ components: its only role is to separate the mass into two elements; the flexibility of the whole transform is otherwise guaranteed by the NF steps with complex deterministic mappings. The interpretation of the figure is otherwise the same.

\subsection{Conditional Density Estimation (CDE) using DIF}

Up to now we have focused on the problem of modeling an unconditional probability distribution, be it for the VI or the VDE problems. 
However,
for scientific and approximate inference purposes in the likelihood-free setting \citep{LikelihoodFree}, 
modeling a conditional pdf $p(x|\omega)$, where $\omega$ is some covariate rv, 
is also a relevant problem. 
It happens that NF can easily be turned into Conditional Density models;
as we now see, 
DIF can also be used for the same purpose. 
In this section we will briefly explain the principles of CDE using NF, and see how the discussion can be extended to DIF.

Recall that a NF is given by a change of variable $T$
with input $z$; therefore in order to obtain a conditional NF, the mapping $T$ must be a function of the covariate $\omega$, such that for fixed $\omega$, $T(.;\omega)$ is a C1-diffeomorphism. This corresponds to defining a conditional transport $\pi(x|z,\omega) = \delta_{T^{-1}(z;\omega)}(x)$,
so the resulting conditional pdf reads $\psi(x|\omega) = q\paren{T(x;\omega)}|\det \mathrm{J}_{T(.;\omega)}(x)|$.
In practice, since the mapping $T$ is classically  parameterized by an NN, this can be achieved for instance by augmenting the input of the NN with the covariate $\omega$ (see for example \citep[\S 3.4]{MAF}).

Now if we relax the hypothesis of an invertible deterministic transport and consider a discrete conditional stochastic transport of the form $\pi(x|z, \omega) = \sum_{k=1}^K w_k(z; \omega) \delta_{T_k^{-1}(z; \omega)}(x)$, then we obtain a conditional DIF model with pdf:
\begin{equation*}
    \psi(x|\omega) = \sum_{k=1}^K w_k\paren{T_k\paren{x;\omega}; \omega} q\paren{T_k\paren{x; \omega}}\abs{\det \mathrm{J}_{T_k\paren{.; \omega}}(x)}
\end{equation*}
Therefore, in order to use a DIF as a CDE model, 
we simply transform $T_k$ and $w_k$ 
(for $k=1,...,K$) 
into functions of the covariate $\omega$. 
For fixed $\omega$,
the associated transform is a DIF
as defined above,
and as such benefits from straightforward sampling and evaluation of the pdf (see section \ref{dif}). 

Let us  propose a parameterization of conditional DIF in the spirit of section \ref{Parameterization}. 
First, the probability functions described in \ref{probabilitypartitioning}
can effortlessly be turned into conditional partitioning functions of the latent space $z$ 
which depend on $\omega$, 
denoted as $w_k(z; \omega)$. Indeed, by augmenting the input $z$ with the covariate $\omega$, the output of the NN is now the vectors of categorical probabilities $\bracket{w_1\paren{z; \omega},..., w_K\paren{z; \omega}}$.
Next C1-diffeomorphisms $T_1,...,T_K$ can be turned into functions of the covariate $\omega$ by simply 
turning the locations and scales into functions of $\omega$. We can use an approach similar to that 
used in Mixture Density Networks (MDN) \citep{MixtureDensityNetworks}, 
where an NN function predicts the location $\mu_k(\omega)$ and log-scales $\log\paren{s_k(\omega)}$ for $k=1,...,K$.

Let us finally consider the optimization objective involved in the CDE problem
(independently of the structure used for the surrogate $\psi(x|\omega)$,
be it a DIF, an NF, an MDN or another model).
We assume that we dispose of samples
$\paren{\omega_i,x_i}$ for $i=1,...,M$,
such that $\omega_i \sim p_\omega(\omega)$ 
(the prior pdf of rv $\omega$)
and $x_i \sim p(x|\omega_i)$, 
but the conditional pdf 
$p(x|\omega)$
cannot be evaluated. We will build the conditional surrogate $\psi(x|\omega)$
of
$p(x|\omega)$
by minimizing the following $\mathrm{D}_{\mathrm{KL}}$:
\begin{equation*}
    \underset{\Psi}{\arg\min}  \mathrm{D}_{\mathrm{KL}}\paren{p_\omega(\omega)p(x|\omega) || p_\omega(\omega)\psi(x|\omega)} =\underset{\Psi}{\arg\max} \E_{p_\omega(\omega)p(x|\omega)}\bracket{\log\paren{\psi(x|\omega)}}.
\end{equation*}
In the end, by using an MC approximation of this expectation
based on the samples at hand, 
the $D_{\mathrm{KL}}$ minimization reduces to maximizing the conditional likelihood:
\begin{equation*}
    \underset{\Psi}{\arg\max}\sum_{i=1}^M \log\paren{\psi(x_i|\omega_i)}.
\end{equation*} 

\section*{Conclusion}

In this paper, we have explored DIF as a methodology to construct parametric surrogates in order to tackle the VI or VDE problems. 
As an extension of NF, DIF produce high flexibility while remaining convenient to use as they are well suited for sampling and density estimation; moreover they do not suffer from the NF topological limitation when targeting pdf with disjoint support. 
On the other hand, DIF also extend mixture density models, and leverage flexible partitioning functions
in order to capture detailed and edged distributions.

\newpage
\appendix

\section{DIF reverse kernel and marginal distribution}
\label{Psi-and-reverse-kernel}

We first check that the function
\begin{align*}
    \overleftarrow{\Pi}: & \mathbb{R}^d \times \mathcal{B}(\mathbb{R}^d) arrow [0, \infty[ \\
    & z, A \rightarrow \overleftarrow{\Pi}(z,A) = \sum_{k=1}^K w_k(z)\mathbbm{1}(T_k^{-1}(z) \in A)
\end{align*}
is a valid transition kernel: 
for fixed $z \in \mathbb{R}^d$, $\overleftarrow{\Pi}(z,.)$ is a probability measure; while for fixed $A \in \mathcal{B}(\mathbb{R}^d)$, $\overleftarrow{\Pi}(.,A)$ is a measurable function. 
On the other hand, 
$\mathbbm{R}^d$ is a Polish space endowed with its Borel $\sigma$-field, 
so the reverse transition Kernel $\overrightarrow{\Pi}$ exists. 
Since for a given $x$, 
only the values $\paren{T_1(x),..., T_K(x)}$ may have produced $x$, 
the reverse transition kernel indeed takes the form:
\begin{align*}
    \overrightarrow{\Pi}: & \mathbb{R}^d \times \mathcal{B}(\mathbb{R}^d) \rightarrow [0, \infty[ \\
    & x, B \rightarrow \overrightarrow{\Pi}(x,B) = \sum_{k=1}^K v_k(x)\mathbbm{1}\paren{T_k(x) \in B},
\end{align*}
in which
$v_k(x) = \mathbbm{P}(Z = T_k^{-1}(\widetilde{X})|\widetilde{X}=x)$. Next, for any  
$A,B\in\mathcal{B}(\mathbbm{R}^d)$, we have
\begin{align}
    \nonumber
    \mathbbm{P}(\widetilde{X} \in A, Z\in B)& = \int_B \overleftarrow{\Pi}(z, A)q(z)\mathrm{d}z \\
    \nonumber
    & = \int_B \paren{\int_A \sum_{k=1}^Kw_k(z)\delta_{T_k^{-1}(z)}(x) \mathrm{d} x} q(z) \mathrm{d}z \\
    \nonumber
    \label{Proba-A-B}
    & =\sum_{k=1}^K \int_{B \cap T_k(A)} w_k(z)q(z)\mathrm{d}z \\
    & =\sum_{k=1}^{K}\int_{T_k^{-1}(B) \cap A}w_k\paren{T_k(x)}q\paren{T_k(x)}\abs{\det \mathrm{J}_{T_k}(x)}\mathrm{d}x .
\end{align}

Two cases are of particular interest:
\begin{itemize}
\item
Set $B = \mathbbm{R}^d$. Since $T_k^{-1}(B) = \mathbbm{R}^d$,
\eqref{Proba-A-B} becomes
\begin{align*}
    \mathbbm{P}(\widetilde{X} \in A) & =
    \sum_{k=1}^K \int_{A} w_k\paren{T_k(x)}q\paren{T_k(x)}\abs{\det \mathrm{J}_{T_k}(x)}\mathrm{d}x \\ 
    &= \int_{A} 
    \underbrace{\sum_{k=1}^K  w_k\paren{T_k(x)}q\paren{T_k(x)}\abs{\det \mathrm{J}_{T_k}(x)}}_{\psi(x)}
    \mathrm{d}x ,
\end{align*}
so 
$\widetilde{X}$ admits pdf $\psi$ wrt Lebesgue measure.
\item
Set $A = \mathbbm{R}^d$. Equation \eqref{Proba-A-B} becomes 
\begin{align*}
    \mathbbm{P}(Z \in B) & = \sum_{k=1}^{K}\int_{T_k^{-1}(B) }w_k\paren{T_k(x)}q\paren{T_k(x)}\abs{\det \mathrm{J}_{T_k}(x)}\mathrm{d}x \\
    & = 
    \sum_{k=1}^{K}\int_{T_k^{-1}(B)}\frac{w_k\paren{T_k(x)}q\paren{T_k(x)}\abs{\det \mathrm{J}_{T_k}(x)}}{\psi(x)}\psi(x)\mathrm{d}x \\
    & =
    \sum_{k=1}^{K}\int_{\mathbbm{R}^d}\frac{w_k\paren{T_k(x)}q\paren{T_k(x)}\abs{\det \mathrm{J}_{T_k}(x)}}{\psi(x)}\mathbbm{1}_{T_k^{-1}(B)}(x)\psi(x)\mathrm{d}x \\
    & =
    \int_{\mathbbm{R}^d}
    \underbrace{\sum_{k=1}^K \frac{w_k\paren{T_k(x)}q\paren{T_k(x)}\abs{\det \mathrm{J}_{T_k}(x)}}{\psi(x)}\mathbbm{1}_{T_k^{-1}(B)}(x)}_{\overrightarrow{\Pi}(x,B)}
    \psi(x)\mathrm{d}x ,
\end{align*}
so $\overrightarrow{\Pi}(x,B) = \sum_{k=1}^K v_k(x)\mathbbm{1}\paren{T_k(x) \in B}$
where $v_k(x)$ is given by \eqref{v_k}.
\end{itemize}

\section{Derivation of GEM objective}\label{GradientEM}
In this section we will explicit model parameters at step $t$ using superscript as in $\psi^{(\theta_t)}(x)$. First, for purpose of conciseness, let us define a subsidiary function which describes the joint pdf of $(\widetilde{X},U)$ for model parameters $\theta$ (see \eqref{DIFDensity}): 
\begin{equation*}
    h_k^{\paren{\theta}}\paren{x} = w_k^{\paren{\theta}}\paren{T_k^{\paren{\theta}}\paren{x}}q\paren{T_k^{\paren{\theta}}\paren{x}}|\det J_{T_k^{\paren{\theta}}}\paren{x}|,\forall{x} \in \mathbbm{R}^d \text{ and } k=1,...,K.
\end{equation*}
Since $\log\paren{\psi^{\paren{\theta}}\paren{x}} =
\E_{\rho}\bracket{ 
\log\paren{\psi^{\paren{\theta}}\paren{x}}}$, where the expectation is taken with respect to discrete categorical rv $U$ with a probability measure $\rho$ such that $U\sim {\rm{Categorical}}\paren{\rho(1),...,\rho(K)}$, we can write:
\begin{align*}
\log\paren{\psi^{\paren{\theta}}\paren{x}} &= 
\mathbbm{E}_{\rho}\bracket{\log\paren{h_U^{\paren{\theta}}\paren{x}} - \log\paren{\mathbbm{P}^{\paren{\theta}}\paren{U|x}}} \\
& = \mathbbm{E}_{\rho}\bracket{\log\paren{h_U^{\paren{\theta}}\paren{x}} - \log\paren{\rho\paren{U}}} \\& \text{    } + \mathbbm{E}_{\rho}\bracket{\log\paren{\rho\paren{U}} - \log\paren{\mathbbm{P}^{\paren{\theta}}\paren{U|x}}}. 
\end{align*}
The last term is $
    D_{\mathrm{KL}}\paren{\rho\paren{U}||\mathbbm{P}^{\paren{\theta}}\paren{U|x}} \geq 0$, hence we have: 
\begin{equation}
    \label{majorationMM}
    \log\paren{\psi^{\paren{\theta}}\paren{x}} \geq \mathbbm{E}_{\rho}\bracket{\log\paren{h_U^{\paren{\theta}}\paren{x}} - \log\paren{\rho\paren{U}}},
\end{equation} 
where equality holds if and only if
\begin{equation}
    \label{egaliteMM}
    \rho\paren{k} = \mathbbm{P}^{\paren{\theta}}\paren{U=k|x}\text{ for all values } k = 1,...,K.
\end{equation}

Let us finally turn to an iterative optimization scheme.
Let 
$\theta_t$
be the current parameter.
Let us set
$\rho\paren{k} =\mathbbm{P}^{\paren{\theta_t}}\paren{U= k|x}= \mathbbm{P}^{\paren{\theta_t}}\paren{z = T_k(x)|x} \stackrel{\eqref{v_k}}{=} v_k^{(\theta_t)}(x)$ for all $k=1,...,K$, 
and let us sum for $x=x_1,..., x_M$. The rhs of \eqref{majorationMM} yields a function $g_{\theta_t}(\theta)$ which reads: 
\begin{align*}
    &g_{\theta_t}\paren{\theta} = \sum_{i=1}^M\sum_{k=1}^K v_k^{(\theta_t)}\paren{x_i}\log\paren{\frac{h_k^{\paren{\theta}}\paren{x_i}}{v_k^{(\theta_t)}\paren{x_i}}},
\end{align*}
and satisfies
\begin{align}
    &g_{\theta_t}\paren{\theta} \stackrel{\eqref{majorationMM}}{\leq} \sum_{i=1}^M\log\paren{\psi^{\paren{\theta}}\paren{x}}\label{majoration}\text{ for all }\theta \in \Theta,\\
    &g_{\theta_t}\paren{\theta_t} \stackrel{\eqref{egaliteMM}}{=} \sum_{i=1}^M \log\paren{\psi^{(\theta_t)}\paren{x_i}}\label{egalité-point-courant} .
\end{align}
Therefore, if we compute $\theta_{t+1}$ via a GA step, 
like that described in Algorithm \ref{gradientEMalgo}
(or, more generally, any method which ensures that $g_{\theta_t}\paren{\theta_{t+1}} \geq g_{\theta_t}\paren{\theta_{t}}$), then
by construction,
we increase the log-likelihood of data $\set{x_1,...,x_M}$ under $\psi$ since:
\begin{equation*}
    \sum_{i=1}^M\log\paren{\psi^{(\theta_{t+1})}\paren{x_i}} \overset{\eqref{majoration}}{\ge} g_{\theta_t}\paren{\theta_{t+1}} \stackrel{GA}{\ge} g_{\theta_t}\paren{\theta_t} \stackrel{\eqref{egalité-point-courant}}{=} \sum_{i=1}^M \log\paren{\psi^{(\theta_{t})}\paren{x_i}}.
\end{equation*}
Finally in our case \eqref{egalité_gradients} reads:
\begin{align*}
    \nabla_{\theta}g_{\theta_t}\paren{\theta}\rvert_{\theta = \theta_t} & = \sum_{i=1}^M\sum_{k=1}^K v_k^{(\theta_t)}(x_i) \nabla_{\theta}\log\paren{h_k^{(\theta)}(x_i)}\rvert_{\theta = \theta_t} \\& = \sum_{i=1}^M\sum_{k=1}^K v_k^{(\theta_t)}(x_i) \frac{\nabla_{\theta}h_k^{(\theta)}(x_i)}{h_k^{(\theta)}(x_i)}\rvert_{\theta = \theta_t} = \sum_{i=1}^M\sum_{k=1}^K v_k^{(\theta_t)}(x_i) \frac{\nabla_{\theta}h_k^{(\theta)}(x_i)\rvert_{\theta = \theta_t}}{h_k^{(\theta_t)}(x_i)} \\ &= \sum_{i=1}^M\frac{1}{\psi^{(\theta_t)}(x_i)} \sum_{k=1}^K \nabla_{\theta}h_k^{(\theta)}(x_i)\rvert_{\theta = \theta_t} = \sum_{i=1}^M\frac{1}{\psi^{(\theta_t)}(x_i)} \nabla_{\theta}\sum_{k=1}^K h_k^{(\theta)}(x_i)\rvert_{\theta = \theta_t} \\& = 
    \sum_{i=1}^M\frac{\nabla_{\theta}\psi^{(\theta)}(x_i)\rvert_{\theta = \theta_t}}{\psi^{(\theta_t)}(x_i)} = \sum_{i=1}^M \nabla_{\theta} \log\paren{\psi^{(\theta)}(x_i)}\rvert_{\theta = \theta_t} = \nabla_{\theta}\sum_{i=1}^M \log\paren{\psi^{(\theta)}(x_i)}\rvert_{\theta = \theta_t} ,
\end{align*}
which validates our construction of functions $\set{g_{\theta_t}}_{t=1,2,...}$ 

\section{Cascading DIF in practice}
\label{cascade-practice}

We now see that the cascaded models discussed in section \ref{mixed_models} can be implemented efficiently for both the VI and VDE problems. We explicit here the according objectives to be optimized for a cascade of two DIF; but with using recursion, this construction can of course be extended to more that two DIF.  
\subsubsection*{VI}

First, we can obtain samples $\widetilde{X}$ from $\Psi$ by sequentially applying $Z_1 \sim \overleftarrow{\Pi}^{[1]}(Z)$ and then $\widetilde{X}\sim \overleftarrow{\Pi}^{[1]}(Z_1)$ to original samples $Z\sim Q$.
This corresponds to the following sampling scheme: 
\begin{align*}
    &\widetilde{X} = {T^{[0]}_{U_0}}^{-1}\paren{{T^{[1]}_{U_1}}^{-1}\paren{Z}} \text{ where } Z\sim Q, U_1\sim {\rm{Categorical}}\set{w_{k_1}^{[1]}(Z)}_{k_1=1,...,K_1} \\& \text{ and } U_0\sim {\rm{Categorical}}\set{w_{k_0}^{[0]}\paren{{T_{U_1}^{[1]}}^{-1}\paren{Z}}}_{k_0=1,...,K_0}.
\end{align*}
As in section \ref{RBVI},
we can use RB in a sequential manner in order to build a differentiable MC approximation of the reverse $D_{\mathrm{KL}}$: 
\begin{equation}\label{doubleRB}
    D_{\mathrm{KL}}\paren{\Psi||P} \approx    \frac{1}{M}\sum_{\stackrel{i=1}{z_i\sim Q}}^M \underbrace{\sum_{k_1 = 1}^{K_1} w_{k_1}^{[1]}(z_i) \underbrace{\sum_{k_0=1}^{K_0} w_{k_0}^{[0]}({T_{k_1}^{[1]}}^{-1}(z_i))\log \paren{\frac{\psi({T_{k_0}^{[0]}}^{-1}({T_{k_1}^{[1]}}^{-1}(z_i)))}{p({T_{k_0}^{[0]}}^{-1}({T_{k_1}^{[1]}}^{-1}(z_i)))}}}_{\E\bracket{J|Z=z_i, U_1}}}_{\E\bracket{J|Z=z_i}}
\end{equation}
which, as explained in section \ref{RBVI}, corresponds to an RB approximation where we successively marginalized out the Categorical latent variables $U_0$ and $U_1$ (compare \eqref{doubleRB} with \eqref{RaoBlackwellDKL}).

\subsubsection*{VDE}
Next, the pdf induced by this cascade model can be easily computed via the following recursion: 
\begin{equation*}
    \psi = \mathcal{F}\paren{\mathcal{F}\paren{q;\overleftarrow{\Pi^{[1]}}};\overleftarrow{\Pi^{[0]}}},
\end{equation*}
hence, as explained in section \ref{DIFDE}, one can use this model for VDE by maximizing the log-likelihood (the equivalent of \eqref{DIFObjective}).
Alternately, as discussed in section \ref{GeneralizedEMApproach}, one can maximize a GEM surrogate, which reads
\begin{align*}
    g_{\theta_t}(\theta) & = \sum_{\stackrel{i=1}{x_i\sim P}}^M \sum_{k_0 = 1}^{K_0} {v_{k_0}^{[0]}}^{(\theta_t)}(x_i) \sum_{k_1 = 1}^{K_1} {v_{k_1}^{[1]}}^{(\theta_t)}({T_{k_0}^{[0]}}^{(\theta_t)}(x_i))\log\paren{\frac{h_{k_0,k_1}^{(\theta)}(x_i)}{{v_{k_0}^{[0]}}^{(\theta_t)}(x_i){v_{k_1}^{[1]}}^{(\theta_t)}({T_{k_0}^{[0]}}^{(\theta_t)}(x_i))}} 
    \end{align*}
    where
    \begin{align*}
    h_{k_0,k_1}(x_i) & = w_{k_0}^{[0]}({T_{k_0}^{[0]}}(x_i))w_{k_1}^{[1]}({T_{k_1}^{[1]}}({T_{k_0}^{[0]}}(x_i)))
    \\
& \times q\paren{{T_{k_1}^{[1]}}\paren{{T_{k_0}^{[0]}}\paren{x_i}}}
    |\det \mathrm{J}_{{T_{k_0}^{[0]}}}(x_i)|.|\det \mathrm{J}_{{T_{k_1}^{[1]}}}\paren{{T_{k_0}^{[0]}}(x_i)}|.
\end{align*}

\vskip 0.2in
\bibliography{bibliography}

\end{document}